\documentclass[final,3p,times,twocolumn]{elsarticle}

\usepackage{xcolor}
\usepackage{hyperref}
\usepackage{amsmath}
\usepackage{amsfonts}
\usepackage[capitalise]{cleveref}
\usepackage{graphicx}
\usepackage{booktabs}
\usepackage{multirow}
\usepackage{amssymb}

\journal{Image and Vision Computing}

\begin{document}

\begin{frontmatter}



\title{Bidirectional Cross-Attention Fusion of High-Resolution RGB and Low-Resolution Hyperspectral Inputs for Multimodal Semantic Segmentation}


\author{Jonas V. Funk$^{a, b, *}$} 
\author{Lukas Roming$^{b}$} 
\author{Andreas Michel$^{b}$} 
\author{Paul B\"acker$^{b}$} 
\author{Georg Maier$^{b}$} 
\author{Thomas L\"angle$^{a,b}$} 
\author{Markus Klute$^{a}$}

\affiliation{organization={$^{a}$KIT, Karlsruhe Institute of Technology},
            postcode={76131 Karlsruhe}, 
            country={Germany}}
            
\affiliation{organization={$^{b}$Fraunhofer IOSB, Fraunhofer Institute of Optronics, System Technologies and Image Exploitation},
            postcode={76131 Karlsruhe}, 
            country={Germany}}

\begin{abstract}
Multimodal semantic segmentation with heterogeneous sensors must reconcile complementary information across modalities that differ in spatial resolution and channel dimensionality. In particular, high-resolution RGB imaging provides detailed spatial structure but often fails to distinguish visually similar materials, whereas hyperspectral imaging (HSI) provides discriminative spectral signatures but at lower spatial resolution. We present Bidirectional Cross-Attention Fusion (BCAF), which aligns high-resolution RGB with low-resolution HSI at their native grids via localized, bidirectional cross-attention, avoiding pre-upsampling or early spectral collapse. BCAF uses two independent backbones: a standard Swin Transformer for RGB and an HSI-adapted Swin backbone that preserves spectral structure through 3D tokenization with spectral self-attention. Although our evaluation targets RGB-HSI fusion, BCAF is modality-agnostic and applies to co-registered RGB with lower-resolution, high-channel auxiliary sensors. On the benchmark SpectralWaste dataset, BCAF delivers strong performance, achieving 75.4\% at 55 images/s. We further evaluate a novel industrial dataset: K3I-Cycling (first RGB subset already released on Fordatis). On this dataset, BCAF reaches 62.3\% mIoU for material segmentation (paper, metal, plastic, etc.) and 66.2\% mIoU for plastic-type segmentation (PET, PP, HDPE, LDPE, PS, etc.). These results show that preserving native-grid spatial detail and spectral structure improves multimodal segmentation under real-time constraints. Code and model checkpoints are publicly available at 
\url{https://github.com/jonasvilhofunk/BCAF_2026}.
\end{abstract}


\end{frontmatter}

\section{Introduction}
\label{sec:Introduction}
Multimodal semantic segmentation with heterogeneous sensors is an important problem in computer vision, particularly when complementary modalities differ in both spatial resolution and channel dimensionality. A common setting combines high-resolution RGB with lower-resolution, high-channel auxiliary sensors, requiring fusion mechanisms that preserve fine spatial detail while exploiting modality-specific information. Automated waste sorting is one instance of this broader problem. In this setting, sensor-based sorting systems operate on high-throughput conveyor belts, where objects must be detected and ejected with compressed air nozzles. This demands reliable, real-time, pixel-level semantic segmentation that can cope with clutter, occlusions, contamination, and variable object appearances \cite{Maier2024}.

A range of sensors is deployed for such sorting tasks. RGB cameras capture color, texture, and shape cues. Hyperspectral imaging (HSI) in the near-infrared (NIR, $\sim$700-1000 nm) and short-wave infrared (SWIR, $\sim$1000-2500 nm) ranges supports material identification. Other application-specific sensors include X-ray transmission, fluorescence systems, laser-induced breakdown spectroscopy (LIBS), and Raman spectroscopy. For polymer sorting in particular, HSI exploits characteristic absorption features to discriminate polyethylene terephthalate (PET), polyethylene (PE), polypropylene (PP), and other polymers \cite{Chang2003, Burns2007}.

Previous work has shown that segmentation performance can often be improved by fusing heterogeneous sensors \cite{zhang2023cmx, zhou2022canet, li2025hybrid, bihler2023multi}. In this article, we follow this direction by fusing RGB and HSI: RGB excels at high-resolution spatial detail but cannot reliably separate materials with similar colors, whereas HSI provides the spectral richness needed for material discrimination but sacrifices spatial resolution. We argue that exploiting the strengths of both modalities without collapsing spectral information or eroding spatial detail is a promising approach for improved semantic segmentation in multimodal RGB+X settings, and it motivates fusion mechanisms that respect each modality’s native resolution and structure while remaining efficient enough for deployment on commodity hardware.

To address this, we introduce Bidirectional Cross-Attention Fusion (BCAF), a multimodal architecture that fuses RGB and HSI without sacrificing either modality’s native strengths. BCAF aligns modalities at their native resolutions across multiple scales via localized bidirectional cross-attention between fine-grid RGB features and coarse-grid multi-slice HSI features. The HSI pathway preserves spectral information, while the RGB pathway retains high-resolution spatial context. Both backbones build upon hierarchical Swin Transformer encoders \cite{Liu2021}. We additionally construct unimodal RGB and HSI segmentation pipelines based on these backbones paired with a shared U-Net-like decoder \cite{Ronneberger2015} and compare them against BCAF. 

This native-resolution, channel-preserving fusion setting represents a broader class of multimodal systems that combine high-resolution RGB with lower-resolution, high-channel auxiliary sensors. Such configurations are common because silicon CMOS processes enable inexpensive, high-resolution RGB cameras, whereas hyperspectral and other non-RGB sensors still require more complex materials and fabrication, which limits their attainable resolution. We therefore expect BCAF to transfer to other RGB+X sensor combinations of this kind.

We evaluate our method on two datasets: SpectralWaste \cite{casao2024spectralwaste} and a novel dataset called K3I-Cycling. We consider two tasks: material-type segmentation (paper, metal, plastic, etc.) and plastic-type segmentation (PET, PP, HDPE, LDPE, PS, etc.), targeting the increasing recycling requirements for packaging waste set by, for example, the European Union \cite{european2025}.

Our study yields five main findings: (i) for both evaluated datasets, increasing RGB input resolution ($256 \rightarrow 512 \rightarrow 1024$) improves Swin-based semantic segmentation, (ii) preserving native high-resolution RGB detail (\emph{resolution as information}) yields gains beyond scaling alone, (iii) the optimal number of HSI spectral slices depends on task granularity: material segmentation peaks with fewer slices, whereas fine-grained plastic-type segmentation benefits from more slices to capture subtle spectral cues, (iv) the adapted HSI backbone preserves more discriminative spectral information than an HSI-to-RGB band-projection baseline using an RGB backbone, and (v) BCAF consistently outperforms unimodal baselines and learned-logit late fusion, achieving real-time throughput. In summary, BCAF delivers strong performance on SpectralWaste, achieving 75.4\% at 55 images/s on an NVIDIA GeForce RTX~4090. On the novel K3I-Cycling dataset, BCAF attains 62.3\% mIoU for material segmentation and 66.2\% mIoU for plastic-type segmentation.

The main contributions are:
\begin{itemize}
  \item A \textbf{native-grid bidirectional fusion} mechanism that aligns multi-scale fine-grid RGB with coarse multi-slice HSI via localized cross-attention, avoiding strong RGB downsampling or PCA-based spectral collapse that is common in existing RGB-X and RGB-HSI fusion methods.
  \item An \textbf{HSI-adapted Swin backbone} with grouped 3D tokenization and factorized spatial–spectral attention, preserving spectral structure throughout all encoder stages instead of projecting HSI to a few pseudo-RGB channels before feature extraction.
  \item A \textbf{systematic study of RGB input resolution and HSI slice count} on SpectralWaste and K3I-Cycling under a unified training and evaluation protocol, quantifying ``resolution-as-information'' gains and task-dependent optimal HSI processing, showing that the proposed BCAF achieves strong performance at real-time throughput.
\end{itemize}

The article is organized as follows:
\Cref{sec:related_work} reviews unimodal RGB/HSI segmentation and multimodal fusion. \Cref{sec:methods} presents BCAF and its constituent modules, including the adapted HSI Swin Transformer. \Cref{sec:experimental_setup} details the datasets (SpectralWaste, K3I-Cycling), training and evaluation protocols. \Cref{sec:results} reports unimodal and fusion results, efficiency, and ablation studies. \Cref{sec:conclusion} concludes, while the Appendix provides additional visualizations and details.

\section{Related Work}
\label{sec:related_work}
We review RGB feature extraction backbones, hyperspectral adaptations of these backbones, semantic segmentation architectures for RGB and HSI, and multimodal fusion strategies, and we identify the gap our method addresses.

\subsection{Feature Extraction Backbones (RGB)}
Early deep backbones for vision were convolutional neural networks (CNNs), which remain competitive for dense prediction under real-time constraints. Lightweight CNNs such as ENet \cite{paszke2016enet} and ICNet \cite{zhao2018icnet} are widely used as efficient feature extractors in industrial settings.

More recent work has adopted hierarchical Vision Transformer backbones tailored for dense prediction. Mixed Transformer (MiT, used in SegFormer) \cite{Xie2021} and Swin Transformer \cite{Liu2021} produce four-stage feature pyramids that are well suited to multi-scale decoding. MiT employs spatial-reduction attention to reduce complexity and underpins SegFormer-B0, which has been shown to enable real-time processing on waste-sorting data \cite{casao2024spectralwaste}. Swin uses shifted-window attention with complexity $O(n \cdot M^2)$, which scales linearly with the number of tokens for a fixed window size $M$. Off-the-shelf ImageNet pretraining and strong downstream performance have made Swin and MiT practical backbones for industrial deployments, including waste sorting \cite{senanayake2025automated, li2025hybrid}. 

In this work, we adopt a Swin encoder as our RGB feature extractor.

\subsection{HSI Adaptations of Feature Extraction Backbones}
Hyperspectral imaging (HSI) provides strong spectral discrimination thanks to material-specific signatures, but its high channel count and often lower spatial resolution pose architectural and efficiency challenges. CNN-based HSI models include 1D spectral CNNs, 3D CNNs, and factorized 2D plus 1D designs. 3D CNNs capture joint spatial-spectral context but are computationally heavy, motivating factorized or dimensionality-reduced approaches under real-time constraints \cite{Ahmad2025}.

Transformer-based HSI backbones adapt image Transformers to handle spectral structure. SpectralFormer \cite{Hong2021} builds on the ViT design \cite{dosovitskiy2021}, using patch embeddings and a Transformer encoder while applying self-attention along the spectral dimension to model band-wise dependencies. Subsequent variants incorporate spatial attention \cite{Yang2022} or factorized spatial-spectral attention \cite{He2021, Sun2022} to jointly capture spatial and spectral context. However, many of these models are designed for patch-wise classification or small tiles rather than dense, real-time segmentation.

In our work, we adapt the hierarchical Swin backbone to HSI by grouping spectral bands into slices and applying factorized spatial-spectral attention, preserving spectral structure without early collapse while remaining compatible with multi-scale decoding.

\subsection{Semantic Segmentation for RGB and HSI}
Semantic segmentation networks map backbone features to pixel-wise predictions. Fully Convolutional Network (FCN) architectures \cite{long2015fully} and U-Net \cite{Ronneberger2015} popularized encoder-decoder designs with skip connections that recover spatial detail while aggregating semantic context. U-Net-like decoders remain standard due to their simplicity, efficiency, and strong performance across modalities.

For RGB imagery, subsequent architectures such as DeepLab \cite{chen2017deeplab} variants and Transformer-based decoders further improved segmentation quality, while lightweight designs like ENet \cite{paszke2016enet}, ICNet~\cite{zhao2018icnet}, and SegFormer-B0 \cite{Xie2021, casao2024spectralwaste} target real-time inference. In industrial waste sorting, encoder-decoder models with hierarchical Transformer backbones (MiT, Swin) and U-Net-like decoders have been evaluated for multimaterial segmentation under throughput constraints~\cite{casao2024spectralwaste, li2025hybrid, senanayake2025automated}. These works demonstrate that real-time semantic segmentation on conveyor belts is feasible, but they also highlight the limitations of RGB-only cues for distinguishing visually similar materials, such as different plastic types \cite{Ji2025}.

For HSI, many models are formulated as patch classification or small-tile segmentation, and often focus on remote-sensing benchmarks \cite{Hong2021, Yang2022, He2021, Sun2022, Ahmad2025}. When applied to dense segmentation, HSI is frequently processed with standard 2D backbones after early spectral collapse into a few bands or pseudo-RGB channels \cite{casao2024spectralwaste, li2025hybrid, Ji2025, senanayake2025automated}, which limits the exploitation of fine-grained spectral information.

In this work, we follow the encoder-decoder paradigm and employ a shared U-Net-like decoder on top of Swin encoders for unimodal RGB, HSI, and their fusion.

\subsection{Multimodal Fusion for Semantic Segmentation}
Fusion strategies are commonly grouped into early, mid-level, and late fusion. Early fusion stacks channels across modalities, often requiring dimensionality reduction (e.g., principal component analysis, PCA), which can discard spectral structure. Mid-level fusion combines intermediate features via concatenation, summation, gating, or cross-attention, enabling richer cross-modal interactions at the cost of alignment complexity and compute. Late fusion merges decisions (e.g., logits), which is efficient but limits feature sharing.

Recent segmentation systems use feature-level cross-/co-attention and multi-scale fusion to align heterogeneous cues. RGB-X methods such as CMX \cite{zhang2023cmx} and CANet \cite{zhou2022canet} rely on up/downsampling to obtain matched spatial grids before fusion, and use cross-attention to exchange information between modalities. In industrial contexts, multimodal setups (e.g., RGB+NIR, RGB+depth) have shown that fusion can outperform unimodal baselines \cite{bihler2023multi, li2025hybrid}.

For the modality combination of RGB and HSI, much of the fusion literature targets remote sensing, with fewer works addressing material discrimination and recycling. Ji et al. \cite{Ji2025} apply an unaltered Swin Transformer backbone to plastic-flake recognition, using learned spectral channel selection and 2D convolutions in the embedding layer to collapse 224 bands and process data purely in the spatial domain. FusionSort \cite{ali2025fusionsort} achieves strong performance on RGB-HSI/multispectral fusion by applying PCA to reduce the spectral dimension prior to fusion. Li et al. \cite{li2025hybrid} propose the Hybrid Long-Range Feature Fusion Network (HLRFF-Net) and benchmark a broad set of state-of-the-art architectures for multimodal waste segmentation, but also rely on PCA-based band reduction (to three components) and resampling operations during fusion.

\subsection{Summary and Gap}
In summary, RGB-based segmentation methods are efficient but struggle to reliably distinguish materials with similar appearance, especially different polymer types with overlapping color and texture cues~\cite{Ji2025}. HSI provides strong spectral cues for material discrimination~\cite{Chang2003, Burns2007, Hong2021} but typically has lower spatial resolution and higher computational cost than RGB sensors~\cite{Ahmad2025}. Moreover, many HSI networks are tailored to patch-wise classification rather than dense, real-time segmentation~\cite{Hong2021, Yang2022, He2021, Sun2022}.

In multimodal settings, prior RGB–HSI fusion methods frequently rely on spectral dimensionality reduction (e.g., PCA) and operate mainly in the spatial domain, collapsing the hyperspectral cube into a few pseudo-RGB channels and limiting the exploitation of fine-grained spectral structure~\cite{ali2025fusionsort, li2025hybrid}. Furthermore, several architectures resize RGB data to the lower HSI resolution before fusion~\cite{li2025hybrid}, discarding high-frequency spatial information that is crucial for segmenting small or thin objects on conveyor belts. Together, these choices leave a gap for multimodal architectures that preserve spectral structure, respect each modality’s native resolution, and still meet real-time constraints on industrial conveyor-belt data.

BCAF is designed explicitly to address this gap. Compared to existing RGB–X and RGB–HSI fusion approaches, it follows a different design philosophy. CMX~\cite{zhang2023cmx} assumes similar spatial resolution across modalities and performs cross-modal attention on matched grids obtained by up/downsampling, whereas BCAF fuses RGB and HSI directly on their native grids without degrading the high-resolution RGB stream. FusionSort~\cite{ali2025fusionsort} and HLRFF-Net~\cite{li2025hybrid} achieve strong performance on multimodal waste segmentation, but both rely on PCA-based band reduction to a few components and early resampling, which collapses fine-grained hyperspectral structure and reduces effective RGB resolution. In contrast, BCAF preserves the spectral axis throughout the HSI backbone via grouped 3D tokenization and spectral self-attention, then performs localized bidirectional cross-attention between fine-grid RGB features and coarse multi-slice HSI features before a lightweight spectral pooling and gating stage.

\section{Methods}
\label{sec:methods}
We first present an overview of the proposed bidirectional spectral cross-attention fusion (BCAF) architecture. In subsequent subsections, we detail the modality-specific RGB and HSI backbones, the multimodal fusion mechanism and the shared segmentation head.

\subsection{Architecture Overview}
\begin{figure*}[t]
  \centering
  \includegraphics[width=1.0\linewidth,trim={0.0cm 1.0cm 0.0cm 0.0cm},clip]{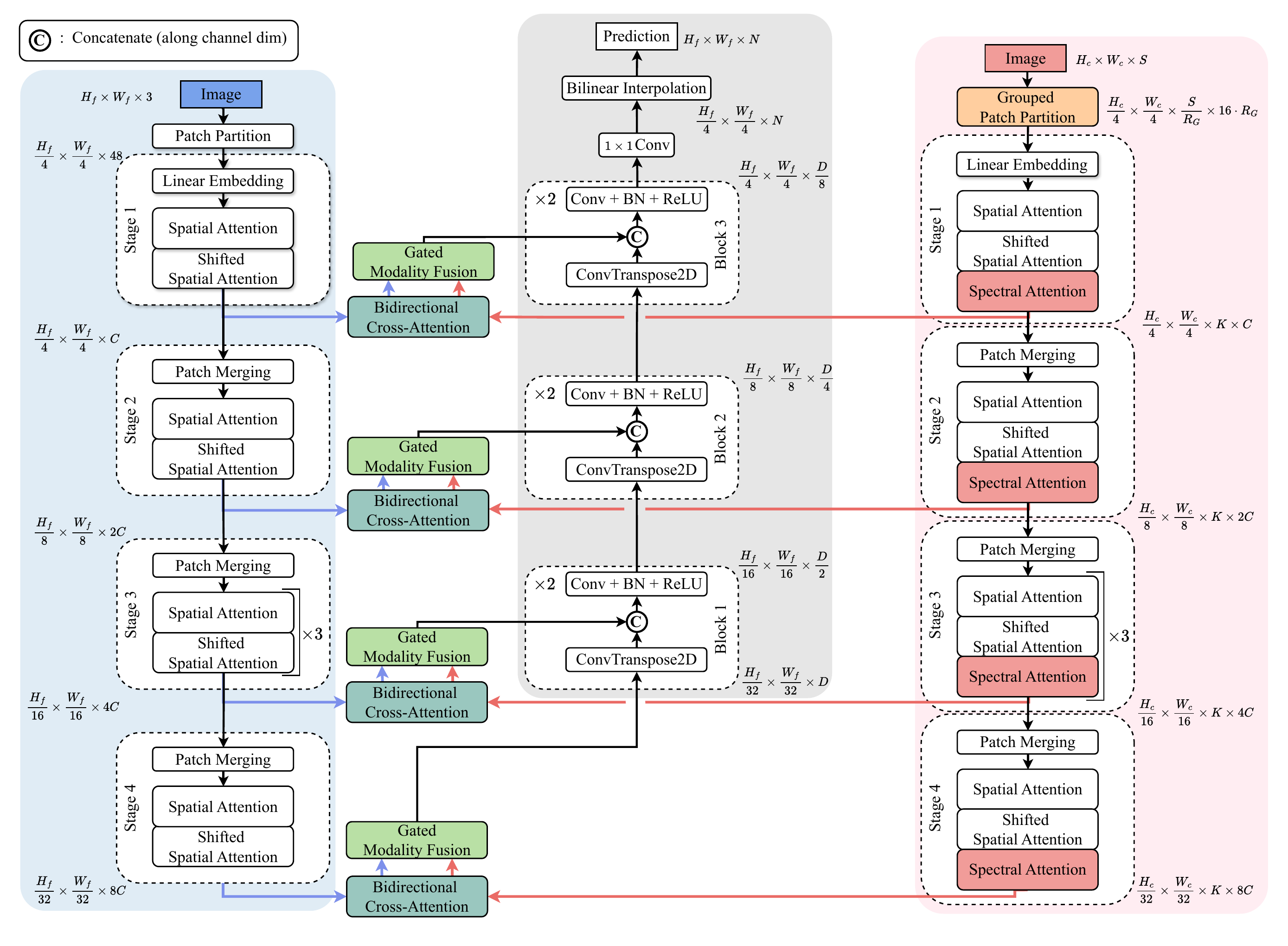}
\caption{BCAF architecture. Left: RGB backbone (blue). Middle: shared decoder/segmentation head (grey). Right: HSI backbone (red) with bidirectional spectral cross-attention and gated modality fusion modules (green).}
  \label{fig:Architecture}
\end{figure*}

\Cref{fig:Architecture} shows the BCAF architecture for semantic segmentation of co-registered RGB-HSI scenes that may differ in spatial resolution. BCAF comprises an RGB backbone (blue), an HSI backbone (red), bidirectional spectral cross-attention (dark green), a gated modality-fusion module (light green), and a shared U\mbox{-}Net-like decoder/segmentation head (grey). Modality-specific backbones enable unimodal training and reuse, while the shared decoder ensures comparable capacity across RGB-only, HSI-only, and fused models.

BCAF is designed around three requirements: (i) preserve high-resolution RGB detail, (ii) preserve HSI spectral structure without early collapse to a few bands, and (iii) remain efficient enough for real-time deployment on conveyor belts. To this end, both backbones follow a multi-scale encoder-decoder paradigm, and fusion is performed locally at the feature level via bidirectional cross-attention at the native grids of each modality.

\textbf{Encoder.}
Both backbones follow a four-stage hierarchical Swin Transformer design with patch partition and patch merging. The RGB backbone performs spatial self-attention in shifted windows and acts as a pure spatial feature extractor for inputs $H_f \times W_f \times 3$. The adapted HSI Swin Transformer preserves the hyperspectral axis for inputs $H_c \times W_c \times S$ by grouping spectral bands into $K$ slices and applying a factorized spatial-spectral attention design to model spectral structure without early dimensionality reduction (e.g., PCA). The spatial ratio $r$ satisfies $H_f=rH_c$ and $W_f=rW_c$ and remains constant across stages because both encoders downsample equally.

\textbf{Fusion and decoder.}
Bidirectional local cross-attention aligns fine-grid RGB features with coarse-grid HSI features at native resolutions, followed by gated modality fusion. Unless otherwise stated, these modules are applied at all four stages. A single shared 2D decoder consumes the fused features via standard skip connections across stages, providing a lightweight architecture that reuses existing encoder-decoder designs.

\textbf{Notation.}
We use channels-last tensor shapes. RGB features are $H_f \times W_f \times C$ (fine grid), and HSI features are $H_c \times W_c \times K \times C$ (coarse grid with $K$ spectral slices). $S$ denotes the number of raw HSI bands. $K$ denotes the slice count after grouping. $r$ denotes the integer spatial ratio with $H_f=rH_c$, $W_f=rW_c$. $C$ denotes the per-stage embedding width. $D$ denotes the decoder width and $N$ denotes the number of classes. Implementation details and hyperparameters are provided in \Cref{subsec:training_and_implementation_details}.

\begin{table}[t]
\centering
\footnotesize
\caption{Notation summary used throughout the methods.}
\label{tab:notation}
\begin{tabular}{ll}
\toprule
Symbol & Meaning \\
\midrule
$H_f,\; W_f$ & Fine-grid (RGB) spatial size (height, width) \\
$H_c,\; W_c$  & Coarse-grid (HSI) spatial size (height, width) \\
$r$      & Spatial ratio: $H_f = r\,H_c,\; W_f = r\,W_c$ \\
$K$    & Number of HSI spectral slices retained after grouping \\
$S$              & Number of raw HSI bands \\
$C$       & Channel width per encoder stage  \\
$D$       & Decoder widths per upsampling stage\\
$N$          & Number of foreground classes \\
\bottomrule
\end{tabular}
\end{table}

\subsection{RGB Backbone}
\label{subsec:RGB_Backbone}
The RGB backbone provides a strong high-resolution spatial feature extractor that captures color, texture, and shape cues from the RGB image while remaining efficient enough for real-time inference. We adopt a hierarchical Swin Transformer encoder~\cite{Liu2021} as the RGB backbone because its shifted-window self-attention scales linearly with the number of tokens for a fixed window size, which allows us to increase RGB input resolution under practical compute budgets. Its four-stage feature pyramid aligns naturally with our multi-scale fusion and U-Net-like decoder, and ImageNet-pretrained Swin models have demonstrated strong performance in industrial dense prediction tasks, including waste sorting. The encoder outputs feature maps at $\{1/4, 1/8, 1/16, 1/32\}$ of the input resolution, which are consumed by the shared decoder.

\subsection{HSI Backbone}
\label{subsec:HSI_Backbone}
The HSI backbone is designed to preserve and model the rich spectral structure of hyperspectral inputs across all encoder stages, so that material-discriminative information is available for fusion rather than being collapsed at the input. We adapt Swin Transformer to hyperspectral inputs with a factorized 2D+1D attention design that preserves spectral structure and supports unimodal HSI segmentation across variable spectral slice counts.

\textbf{Grouped patch partition.}
Let $S$ be the number of raw HSI bands. We replace the RGB patch partition over $(4{\times}4{\times}3)$ with a grouped partition over $(4{\times}4{\times}R_G)$ bands. Each $(4{\times}4{\times}R_G)$ cube is projected by a convolution into one token, yielding a 3D token lattice of size $(H_c/4)\times (W_c/4)\times K$ with $C$ channels, with $K=\lceil S/R_G\rceil$ spectral slices with exact padding, if needed (see \Cref{subsec:training_and_implementation_details}). We refer to this backbone configuration as HSI-$K$.

\begin{figure*}[t]
  \centering
\includegraphics[width=1.0\linewidth,trim={0.0cm 0.3cm 0.0cm 0.0cm},clip]{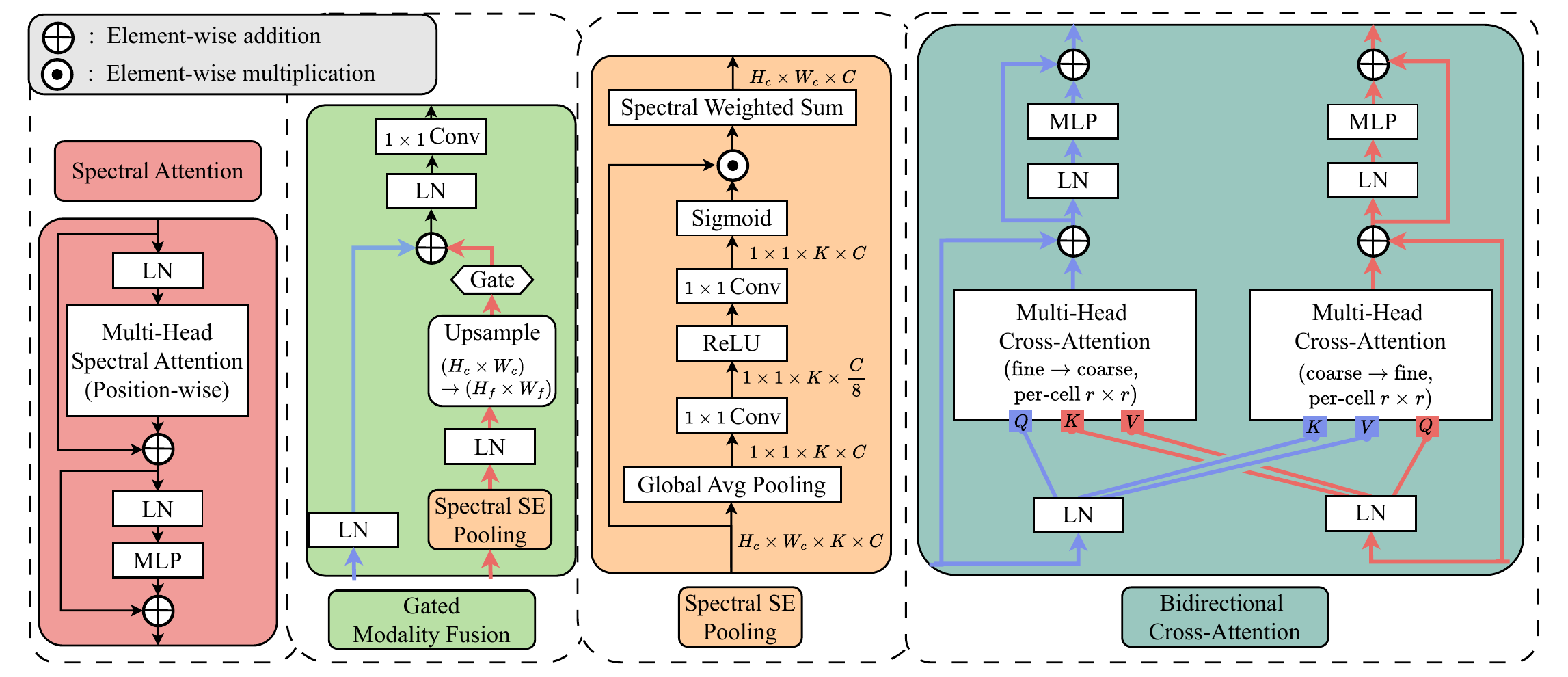}
  \caption{Module details. Left (red): position-wise spectral self-attention. Middle/left (light-green): gated modality fusion. Middle/right (orange): spectral SE Pooling module. Right (dark-green): bidirectional cross-attention modules.}
    \label{fig:Modules}
\end{figure*}

\textbf{Factorized attention ($K>1$).}
At each stage, we apply standard Swin spatial window attention (and its shifted variant) independently to every spectral slice, followed by spectral multi-head self-attention along the slice axis ($K$) at each spatial location. A learnable absolute spectral positional embedding provides ordering information for the $K$ slices. The number of attention heads follows Swin Transformer defaults. This factorization preserves spectral structure while decoupling spatial and spectral modeling. A block-level view of the spectral attention appears in \Cref{fig:Modules} (left, red).

\textbf{Spatial-only patch merging.}
Patch merging aggregates tokens only across spatial dimensions, halving spatial resolution at each stage while keeping the slice count $K$ unchanged. Consequently, spectral resolution remains consistent across all four stages.

\textbf{Outcome.}
For $K{>}1$, the adapted Swin Transformer backbone processes HSI cubes without early dimensionality reduction (e.g., PCA) and produces hierarchical 3D feature maps (height $\times$ width $\times K$ slices) at four spatial resolutions for subsequent semantic segmentation.

\textbf{HSI-1 (spectral collapse, 2D backbone).}
When $K{=}1$, there is a single spectral slice and the pathway is purely 2D. We embed the $S$-band input with a learned $4{\times}4{\times}S \rightarrow C$ convolution and process it with a standard 2D Swin Transformer backbone and the shared 2D decoder. No spectral self-attention is applied. This baseline collapses the spectral axis at input and serves as a strong 2D reference for comparisons to $K{>}1$ settings that preserve spectral structure.

\subsection{Spectral SE Pooling Module}
\label{subsec:Spectral_SE_Pooling_Module}
The spectral squeeze-excitation (SE) pooling module compresses the $K$ HSI slices at each spatial location into a single feature map in a data-adaptive way, allowing a standard 2D decoder to be reused while retaining the most informative spectral contributions. The design is inspired by SENet \cite{hu2018squeeze} and operates along the spectral dimension. Its block-level architecture is shown in \Cref{fig:Modules} (Middle/right, orange).

\textbf{Formulation.}
Given HSI input features $\mathbf{F_\mathrm{hsi}} \in \mathbb{R}^{H \times W \times K \times C}$, we first perform spatial global average pooling:
\[
\mathbf{z} = \mathrm{AvgPool}_{H,W}(\mathbf{F_\mathrm{hsi}}) \in \mathbb{R}^{K \times C}.
\]
Then we compute per-slice, per-channel gates with a two-layer bottleneck (applied independently for each $k$):
\[
\mathbf{a} = \sigma\ \big(f_2(\mathrm{ReLU}(f_1(\mathbf{z})))\big) \in \mathbb{R}^{K \times C},
\]
where $f_1: \mathbb{R}^{C}\!\to\ \mathbb{R}^{C_h}$ and $f_2: \mathbb{R}^{C_h}\ \to\ \mathbb{R}^{C}$ are pointwise ($1{\times}1$) projections shared across $k$, with $C_h=\max(\lfloor C/8 \rfloor, 8)$. Finally, we aggregate across the spectral dimension:
\[
\hat{\mathbf{F}}_{\mathrm{hsi}} = \sum_{k=1}^{K} \mathbf{a}_k \odot \mathbf{F}_{\mathrm{hsi},k} \;\in\; \mathbb{R}^{H \times W \times C},
\]
where $\mathbf{a}_k \in \mathbb{R}^{1 \times 1 \times C}$ is the gate for slice $k$, 
$\mathbf{F}_{\mathrm{hsi},k} \in \mathbb{R}^{H \times W \times C}$ is the $k$‑th spectral slice,
and $\odot$ denotes channel-wise multiplication broadcast over $H \times W$. This spectral-to-spatial reduction produces a 3D feature map suitable for 2D decoders.

\subsection{Bidirectional Cross-Attention}
\label{subsec:Bidirectional_Cross_Attention}
The bidirectional cross-attention block lets each low-resolution HSI cell and its $r\times r$ high-resolution RGB neighbors exchange information locally, so that spectral cues refine RGB features and spatial context refines HSI features without resampling either modality (\Cref{fig:Modules}, dark green). Let RGB features lie on a fine grid $H_f\times W_f$ and HSI features on a coarse grid $H_c\times W_c$, with $H_f=rH_c$ and $W_f=rW_c$ for integer $r\ge 1$. Each coarse HSI location $(u,v)$ has $K$ spectral slices and $r^2$ co-registered RGB “children”. Attention is computed strictly within each coarse location $(u,v)$ (no mixing across parents). Multi-head projections are used in both directions, where the per-head dimension is $d_h=C/h$. Per-head indices are omitted for clarity. At each stage we have the two sets of features:
\[
\mathbf{F_{\mathrm{rgb}}}\in\mathbb{R}^{H_f\times W_f\times C},\qquad
\mathbf{F_{\mathrm{hsi}}}\in\mathbb{R}^{H_c\times W_c\times K\times C}.
\]

\textbf{Fine$\to$Coarse (RGB queries HSI).}
For each HSI parent $(u,v)$, we extract $r^2$ RGB children $X_{\mathrm{rgb}}(u,v)\in\mathbb{R}^{r^2\times C}$ via pixel-unshuffle \cite{shi2016real} (space-to-depth), which rearranges every $r\times r$ fine-grid neighborhood into a length-$r^2$ vector per $(u,v)$. The $K$ HSI slices are $X_{\mathrm{hsi}}(u,v)\in\mathbb{R}^{K\times C}$. With linear projections for queries, keys, and values,
\[
Q_{\mathrm{rgb}}=X_{\mathrm{rgb}}W_Q^{r\to h},\quad
K_{\mathrm{hsi}}=X_{\mathrm{hsi}}W_K^{r\to h},\quad
V_{\mathrm{hsi}}=X_{\mathrm{hsi}}W_V^{r\to h},
\]
the local scaled dot-product cross-attention at $(u,v)$ is
\[
O^{\mathrm{f\to c}}=\mathrm{softmax}\ \Big(\frac{Q_{\mathrm{rgb}}K_{\mathrm{hsi}}^\top}{\sqrt{d_h}}\Big)V_{\mathrm{hsi}}
\;\in\;\mathbb{R}^{r^2\times C},
\]
with softmax over the $K$ keys. Outputs $O^{\mathrm{f\to c}}$ are reassembled to the fine grid by pixel-shuffle (depth-to-space), linearly projected, fused with the RGB features via a residual connection, and passed through a position-wise FFN, preserving $\mathbb{R}^{H_f\times W_f\times C}$.

\textbf{Coarse$\to$Fine (HSI queries RGB).}
Symmetrically, at each parent $(u,v)$ the $K$ HSI slices attend to the $r^2$ RGB children:
\[
Q_{\mathrm{hsi}}=X_{\mathrm{hsi}}W_Q^{h\to r},\quad
K_{\mathrm{rgb}}=X_{\mathrm{rgb}}W_K^{h\to r},\quad
V_{\mathrm{rgb}}=X_{\mathrm{rgb}}W_V^{h\to r},
\]
\[
O^{\mathrm{c\to f}}=\mathrm{softmax}\ \Big(\frac{Q_{\mathrm{hsi}}K_{\mathrm{rgb}}^\top}{\sqrt{d_h}}\Big)V_{\mathrm{rgb}}
\;\in\;\mathbb{R}^{K\times C},
\]
with softmax over the $r^2$ keys. These updates are added residually to the HSI slices and passed through a position-wise FFN, preserving $\mathbb{R}^{H_c\times W_c\times K\times C}$.

\textbf{Special case $r{=}1$.}
When grids coincide, RGB$\to$HSI reduces to $1{\times}K$ attention and HSI$\to$RGB reduces to single-key attention at each location.

\subsection{Modality Fusion Module}
\label{subsec:Modality_Fusion_Module}
The modality-fusion module merges the RGB and HSI branches into a single 2D feature map per stage for the shared decoder. It first applies spectral SE pooling (\Cref{subsec:Spectral_SE_Pooling_Module}) to compress the HSI slice axis while preserving the channel dimension, then upsamples the result by $r$ (nearest-neighbor) to match the RGB grid, yielding $\mathbf{\hat F_{\mathrm{hsi}}} $ (identity when $r{=}1$). Both branches are LayerNorm (LN) normalized. A learned per-channel gate $\boldsymbol{\alpha}\!\in\!\mathbb{R}^{C}$ (sigmoid-activated) modulates the HSI contribution, and the fused output is
\[
\mathbf{F_{\mathrm{fuse}}} = \mathrm{LN}\ \Big(\mathrm{LN}(F_{\mathrm{rgb}}) + \sigma(\boldsymbol{\alpha}) \odot \mathrm{LN}(\hat{F}_{\mathrm{hsi}})\Big)
\;\in\;\mathbb{R}^{H_f\times W_f\times C},
\]
where $\sigma(\cdot)$ denotes the sigmoid and $\odot$ denotes channel-wise multiplication broadcast over $H_f\times W_f$. The module is applied at all four stages after cross-attention, introduces minimal overhead ($O(H_fW_fC)$), preserves spatial resolution, and allows adaptive control of HSI influence relative to RGB. 

\subsection{Segmentation Decoder}
\label{subsec:Segmentation_Head}
The segmentation decoder converts the multi-scale fused (or unimodal) feature maps into dense pixel-wise class predictions, reusing a lightweight U-Net-like structure that is identical for RGB-only, HSI-only, and fused inputs. The decoder expects 2D feature maps at four hierarchical stages and is identical in all cases. Features from the backbones are first channel-aligned via $1{\times}1$ adapters to a common width $D$ per stage. Starting from the deepest stage, the decoder performs three upsampling steps (transpose-convolution with stride $2$), each followed by Conv-BatchNorm-ReLU refinement. Lateral skip connections from shallower stages are concatenated at matching resolutions to preserve fine detail.

A final $1{\times}1$ classifier maps to $N$ classes. Logits are resized to input resolution by bilinear interpolation if needed. In the HSI-only path, each stage is passed through spectral SE pooling to collapse the slice axis ($K$) into a 2D map prior to channel alignment. In fusion, the fused 2D features are channel-aligned and decoded identically.

\subsection{Losses}
We optimize a weighted sum of cross-entropy and Dice losses:
\begin{equation*}
\mathcal{L} \;=\; 0.5\,\mathcal{L}_{\text{CE}} \;+\; 1.5\,\mathcal{L}_{\text{Dice}}.
\end{equation*}
Let $z \in \mathbb{R}^{H \times W \times (N+1)}$ be the logits over $N$ foreground classes plus background. Let $p=\mathrm{softmax}(z)$ denote class probabilities, and $Y \in \{0,1\}^{H \times W \times (N+1)}$ the one-hot labels.
\begin{equation*}
\mathcal{L}_{\text{CE}}
= -\frac{1}{HW}\sum_{h=1}^{H}\sum_{w=1}^{W}\sum_{n=1}^{N+1}
w_n\, Y_{n,h,w}\,\log p_{n,h,w},
\end{equation*}
with median-frequency class weights computed on the training set (background included) to compensate for pixel-level class imbalance:
\begin{equation*}
w_n=\frac{\mathrm{median}\{f_k \mid f_k>0\}}{f_n+\varepsilon_w},\qquad \varepsilon_w=10^{-6}.
\end{equation*}
Dice loss is computed on probabilities with $\epsilon=1.0$, averaged uniformly over all $N{+}1$ classes (background included):
\begin{equation*}
\begin{aligned}
\mathrm{Dice}_{n} &= \frac{2\sum_{h,w} p_{n,h,w}\,Y_{n,h,w}+\epsilon}
{\sum_{h,w} p_{n,h,w}+\sum_{h,w} Y_{n,h,w}+\epsilon},\\
\mathcal{L}_{\text{Dice}} &= 1 - \frac{1}{N+1}\sum_{n=1}^{N+1}\mathrm{Dice}_{n}.
\end{aligned}
\end{equation*}

\section{Experimental Setup}
\label{sec:experimental_setup}
We evaluate our approach on two multimodal RGB-HSI datasets: the public SpectralWaste  dataset\cite{casao2024spectralwaste} and our novel K3I-Cycling dataset. K3I-Cycling is evaluated under two label taxonomies: K3I-Material (primary materials) and K3I-Plastic (plastic-type distinctions). We first describe the datasets, splits, and preprocessing, then detail training and implementation, baselines, and evaluation protocols.

\subsection{Datasets}
\label{subsec:Datasets}

\textbf{SpectralWaste.}
This multimodal benchmark was collected in an operational plastic-waste sorting facility, reflecting realistic indoor recycling lines with densely packed, partially occluded, and contaminated items \cite{casao2024spectralwaste}. Each scene includes a conventional RGB image (Teledyne DALSA Linea, initial resolution $1200 \times 1184$) and a near-infrared hyperspectral cube (Specim FX17, initial $600 \times 640 \times 224$), pairing color cues with material-sensitive spectral signatures. The released subset provides 852 non-overlapping images with 2{,}059 instance annotations across six categories: basket, film, filament, video tape, cardboard, and trash bag. We use the official split of 514/167/171 for train/validation/test. The public release offers aligned RGB at $256 \times 256$ and HSI at $256 \times 256 \times 224$.

For our resolution analysis, we keep HSI fixed at $256{\times}256$ and vary only the RGB input resolution. Specifically, we start from the provided $256{\times}256$ RGB images and upsample them to $512{\times}512$, $1024{\times}1024$, and $2048{\times}2048$, yielding a \emph{more pixels, same information} regime where the effective spatial information is unchanged but the number of pixels processed by the network increases.

\begin{figure*}[t]
    \centering
    \includegraphics[width=1.0\linewidth,trim={4.8cm 3.0cm 5.5cm 0.3cm},clip]{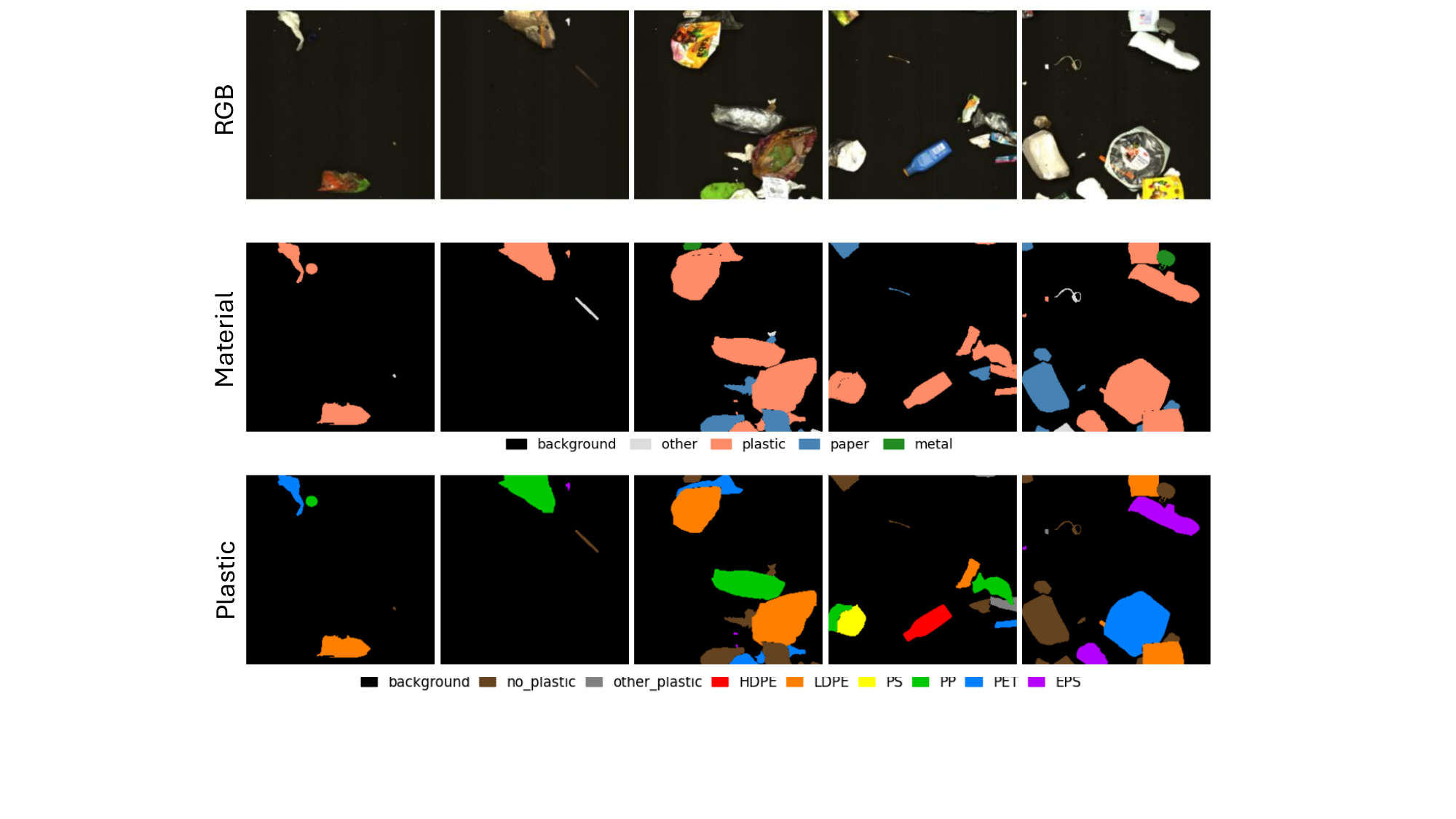}
    \caption{Samples from the K3I-Cycling dataset. Rows (top to bottom): RGB image, labelled K3I-Material masks, labelled K3I-Plastic masks.}
    \label{fig_1}
\end{figure*}

\textbf{K3I-Cycling.}
K3I-Cycling is a lightweight packaging dataset captured with a RGB (SW-4000T-10GE from JAI) and a hyperspectral line-scan camera (FX17e from SPECIM). Samples were transported on a conveyor at $0.2\,\mathrm{m/s}$ with mixed materials and realistic residue/contamination. Items were post-consumer lightweight packaging provided by Lobbe RSW GmbH. Example scenes are shown in \Cref{fig_1}. The dataset contains 354 co-registered RGB-HSI pairs with 4{,}855 labelled object instances. RGB images have $4096 \times 4096$ resolution, HSI cubes have $512 \times 512 \times 205$ with $205$ being the number of spectral channels linearly distributed from $964\,\mathrm{nm}$ to $1668\,\mathrm{nm}$. We adopt a 214/70/70 train/val/test split. All foreground objects are assigned a semantic label. Background corresponds to the conveyor and non-object pixels. We plan to publicly release the RGB images in the future (first subset released), with an expanded set of images and sensors.

RGB and HSI images are co-registered via marker-based calibration. Circular fiducials at the start and end of each of 10 runs are automatically detected. We first rectify the RGB images and correct the nonlinear pixel distribution along the recording line to establish an orthogonal, linearly spaced base coordinate system. A polynomial mapping between RGB and HSI feature coordinates is then fitted, and HSI images are warped into the RGB coordinate system using \texttt{cv2.remap} (OpenCV-Python). 

We evaluate two taxonomies and refer to them as:
\begin{itemize}
  \item \textbf{K3I-Material} (4 primary materials): plastic (2{,}937 instances), paper/cardboard (1{,}072), metal (266), other (580).
  \item \textbf{K3I-Plastic} (8 plastic-related classes): no plastic (1{,}033), LDPE (low-density polyethylene, 813), PP (polypropylene, 765), PET (polyethylene terephthalate, 349), EPS (expanded polystyrene, 184), HDPE (high-density polyethylene, 157), PS (polystyrene, 111), other plastic (1{,}443).
\end{itemize}

Counts refer to instance-level annotations across the full dataset. Mean normalized spectra for all classes are shown in the Appendix (see \Cref{fig:spectra}). We treat 'other' (Material) and 'other plastic' (Plastic) as valid foreground classes and include them in training and evaluation (they count toward $N$). In both taxonomies, 'background' denotes the conveyor belt and non-object pixels. For K3I-Plastic, 'no plastic' is a foreground class for non-plastic objects (e.g., paper/cardboard, metal) and is distinct from background. The 'other plastic' class is used for plastics which are none of the predefined classes or when a plastic item cannot be reliably assigned to one of the listed polymer types (e.g., heavy contamination or uncertain spectra).

\textbf{Input resolutions for K3I-Cycling.}
K3I-Material/Plastic has native RGB resolution $4096{\times}4096$ and HSI resolution $512{\times}512$. We downsample HSI to $256{\times}256$ for all experiments.

For RGB, we consider two regimes:

\begin{enumerate}[(i)]
  \item \textbf{Small-resolution regime} (\emph{more pixels, same information}): we first downsample the RGB image to $256{\times}256$, then upsample this low-resolution image to $256{\times}256$, $512{\times}512$, $1024{\times}1024$, and $2048{\times}2048$. This mirrors the SpectralWaste setting, increasing pixel count without adding spatial information.
  \item \textbf{Native-resolution regime} (\emph{more pixels, more information}): we downsample directly from the native $4096{\times}4096$ to $2048{\times}2048$, $1024{\times}1024$, and $512{\times}512$, allowing us to quantify the impact of retaining higher spatial information on segmentation performance.
\end{enumerate}

\begin{table}[ht]
\centering
\setlength{\tabcolsep}{3pt}
\footnotesize
\caption{Spatial setup used in all experiments. ``Native'' denotes the original resolution, while ``Input'' denotes the resolution to which the data are resized, either by downsampling or upsampling, before being used as model input.}
\label{tab:spatial_setup}
\begin{tabular}{lcc}
\toprule
Dataset / setting      & RGB (Native / Input)              & HSI (Native / Input)  \\
\midrule
SpectralWaste          & 1200$\times$1184 / 256$^2$        & 600$\times$640 / 256$^2$ \\
K3I, scaling-only      & 256$^2$ / 256$^2$–2048$^2$           & 512$^2$ / 256$^2$ \\
K3I, native-resolution & 4096$^2$ / 512$^2$–2048$^2$           & 512$^2$ / 256$^2$ \\
\bottomrule
\end{tabular}
\end{table}

\subsection{Training and Implementation Details}
\label{subsec:training_and_implementation_details}
\textbf{Training regime.}
Training proceeds in two phases with identical optimization hyperparameters: (i) unimodal training of RGB and HSI models (trained independently), and (ii) multimodal fusion fine-tuning that initializes from the unimodal checkpoints and adds bidirectional cross-attention. Each phase runs for 1000 epochs on SpectralWaste and 300 epochs on K3I-Material/Plastic, yielding a comparable number of optimization steps given the smaller size of K3I-Cycling. During fusion, all parameters (both backbones, cross-attention, fusion gates, and decoder) are unfrozen unless noted otherwise.

\textbf{RGB backbone configuration.}
We adopt Swin Transformer Tiny (Swin-T) \cite{Liu2021} as the RGB encoder with patch size $4$, window size $7$, shift $3$, attention depths $(2,2,6,2)$, and heads $(3,6,12,24)$. Embedding dimensions expand across stages $(96,192,384,768)$. DropPath is linearly scheduled to $0.3$. We initialize from \texttt{swin\_tiny\_patch4\_window7\_224}~\cite{rw2019timm}.

\textbf{HSI backbone configuration.}
For the HSI backbone we adapt Swin-T as described in \Cref{subsec:HSI_Backbone}. We initialize from \texttt{swin\_tiny\_patch4\_window7\_224}~\cite{rw2019timm} where shapes match and randomly initialize spectral-specific layers. For example, for adapted Swin-T attention depths of $(3,3,9,3)$, the first $(2,2,6,2)$ blocks per stage are initialized from the pretrained Swin-T weights, while the additional blocks $(1,1,3,1)$ are initialized randomly.

To create the 3D tokenization, we group contiguous bands into $K$ slices. When the raw band count $S\in\{224,205\}$ is not divisible by $K$, we zero-pad to the nearest multiple and set the group size $R_G=S_{\mathrm{pad}}/K$. We adopt a shared-kernel embedding: a single $4{\times}4{\times}R_G\!\to\!C$ convolution is applied to each spectral group with weights shared across groups, producing $(H/4)\times(W/4)\times K$ tokens. For SpectralWaste (224 bands), this yields $R_G=\{224,75,45,32,23\}$ for $K=\{1,3,5,7,10\}$, obtained by padding to $S_{\mathrm{pad}}=225$ for $K\in\{3,5\}$ and to $230$ for $K=10$. For K3I-Material/Plastic (205 bands), this yields $R_G=\{205,69,41,30,21\}$ for $K=\{1,3,5,7,10\}$, obtained by padding to $S_{\mathrm{pad}}=207$ for $K=3$ and to $210$ for $K\in\{7,10\}$.

\textbf{BCAF fusion configuration.}
We use one cross-attention block per stage to limit computational overhead. The number of heads follows the backbones: $(3,6,12,24)$ across the four stages. Cross-attention and fusion parameters are randomly initialized.

\textbf{Shared decoder configuration.}
For the three decoder blocks with channel widths $D=[256,128,64]$ we use dropout $0.0$ for RGB-only models and $0.1$ for HSI-only and fusion models.

\textbf{Data augmentation (train only).}
We apply identical spatial augmentations to RGB and HSI: random rotation ($[-5^\circ,+5^\circ]$), scale ($[0.8,1.3]$), random crop to the target input resolution, and horizontal/vertical flips (each with probability $0.5$). RGB photometric jitter is applied with probability $0.8$ (brightness/contrast/saturation $0.2$, hue $0.03$). HSI spectral jitter uses additive $\mathcal{U}[-0.10,0.10]$ and multiplicative $\mathcal{U}[0.90,1.10]$, applied only to non-zero elements.

\textbf{Background masking for HSI.}
To prevent interpolation bleed of non-zero foreground into the zero-valued HSI background, we construct a binary foreground mask directly from the HSI (non-zero spectrum $=1$, background $=0$), apply the same geometric transforms to the mask using nearest-neighbor, and multiply the transformed mask with the transformed HSI to re-zero the background. This preserves a strictly zero background, avoids bias in downstream normalization, and does not rely on any label information.

\textbf{Normalization.}
RGB uses ImageNet statistics ($\mu_{\text{RGB}}=[0.485,0.456,0.406]$, $\sigma_{\text{RGB}}=[0.229,0.224,0.225]$). HSI uses masked per-channel standardization computed over the training split, excluding background (zeros). For the train split, normalization is applied at the end of data augmentation pipeline.

\textbf{Optimization and schedule.}
We use AdamW (weight decay $0.01$) with target learning rates by parameter group: head $1{\times}10^{-4}$, backbone (pretrained) $1{\times}10^{-5}$, backbone (random init) $1{\times}10^{-4}$. The schedule uses a 5-epoch linear warm-up to the target learning rate (LR), then polynomial decay (power $0.9$) to the end of training. The effective batch size is $8$ via 4-step gradient accumulation (micro-batch $2$ per step). We follow \texttt{timm} defaults and do not apply weight decay to normalization and bias parameters.

\textbf{Seeds and reproducibility.}
We train with five seeds (40-44) on SpectralWaste and three seeds (40-42) on K3I-Cycling, applied to PyTorch, NumPy, data loading, and CUDA. We report mean~$\pm$~std over the corresponding seeds. We enable automatic mixed precision (AMP) and do not use gradient clipping. 

\subsection{Baselines}
\label{subsec:Baselines_Fusion}
\textbf{RGB encoder baselines.}
We compare against SegFormer’s MiT (Mix Transformer) encoders MiT-B0 and MiT-B2 \cite{Xie2021}. To equalize decoder capacity across backbones, we replace SegFormer’s original MLP head with our shared U-Net-like decoder (\Cref{subsec:Segmentation_Head}). Initialization uses ImageNet pretrained MiT weights, and training follows exactly the same data preprocessing, augmentations, optimization, and schedules as for our Swin-T RGB backbone (\Cref{subsec:training_and_implementation_details}).

\textbf{Logit fusion (late-fusion baseline).}
As a simple modality-combination baseline, we fuse unimodal predictions via logit fusion. We load unimodal RGB and HSI checkpoints, obtain their per-pixel logits, bilinearly resize the HSI logits to the RGB grid (\texttt{align\_corners=false}) if needed, concatenate the two along the channel dimension, and apply a learned $1{\times}1$ convolution to produce fused logits. A softmax yields the final probabilities. Training follows exactly the same data preprocessing, augmentations, optimization, and schedules as for our BCAF \Cref{subsec:training_and_implementation_details}. For logit fusion, the unimodal RGB/HSI backbones and the $1{\times}1$ fusion layer are fine-tuned jointly.

\subsection{Metrics and Evaluation Protocol}
\label{subsec:Metric_and_Evaluation_Protocol}
We report per-class Intersection-over-Union (IoU) and mean IoU (mIoU). For each class $c$,
\begin{equation}
\mathrm{IoU}_{n} = \frac{TP_{n}}{TP_{n} + FP_{n} + FN_{n}},
\end{equation}
where $TP_{n}$, $FP_{n}$, and $FN_{n}$ denote the number of true-positive, false-positive, and false-negative pixels for class $n$, accumulated over the evaluation split. The mIoU is the unweighted mean over the evaluated classes $N$, excluding the background class:
\begin{equation}
\mathrm{mIoU} = \frac{1}{|N|} \sum_{n \in N} \mathrm{IoU}_{n}.
\end{equation}

On SpectralWaste (ground-truth masks $256\times256$), model outputs produced at $512\times512$, $1024\times1024$, and $2048\times2048$ are downsampled to $256\times256$ by bilinear interpolation of the raw logits (\texttt{align\_corners=false}), followed by $\arg\max$ over classes, before computing IoU with the original labels to ensure comparability with prior work.

We also assess efficiency and complexity by reporting throughput (images/s) and GFLOPs. Measurements use synthetically generated inputs at the specified resolutions: RGB $256{\times}256{\times}3$, $512{\times}512{\times}3$, $1024{\times}1024{\times}3$, $2048{\times}2048{\times}3$ and HSI $256{\times}256{\times}S$ with $S \in \{224,225,230\}$ depending on spectral padding. Throughput is measured on a single NVIDIA GeForce RTX~4090 with cuDNN autotuning enabled, \texttt{model.eval()}, \texttt{torch.no\_grad()}, batch size $=1$, and FP16 (AMP) disabled (FP32 inference). For each of 20 runs, we perform 100 warm-up forward passes to stabilize kernels, followed by 1000 timed passes using CUDA events with explicit GPU synchronization. We report the median over runs.

We measure GFLOPs with \texttt{fvcore} using matched dummy inputs for unimodal models. For fusion models, encoder GFLOPs are measured with \texttt{fvcore}, while cross-attention, decoder, and adapter layers are computed analytically from feature map dimensions, token dimensions as well as the number of spectral slices $K$.

\section{Results}
\label{sec:results}
This section presents the empirical evaluation of our unimodal RGB and HSI pipelines and our multimodal BCAF. We first analyze unimodal performance to isolate the effects of RGB spatial input resolution and the number of HSI spectral slices $K$ (\Cref{subsec:RGB_effect_of_input_resolution,subsec:HSI_effect_of_spectral_slices}). We then compare BCAF against a late-fusion baseline using learned logit fusion (\Cref{subsec:Results_Fusion}). We further assess computational efficiency  (\Cref{subsec:Results_Computation}), and conduct ablations (\Cref{subsec:Results_Ablation}) on key HSI backbone components (grouped embedding and spectral attention) as well as on fusion-stage placement within BCAF.

As metrics we report class-wise Intersection-over-Union (IoU, \%) and mean IoU (mIoU, \%), alongside images per second (images/s), parameter counts (M), and GFLOPs. Higher is better for IoU/mIoU/images/s ($\uparrow$), and lower is better for Params/GFLOPs ($\downarrow$). SpectralWaste results are averaged over 5 seeds and K3I-Cycling results over 3 seeds (mean~$\pm$~std), under a consistent training protocol for fair comparison.

Qualitative results on SpectralWaste are shown in \Cref{fig_2}, with detailed quantitative results in \Cref{tab:spectralwaste_eval}. For CMX, FusionSort, and HLRFF‑Net we report the mIoU values published in their original works on SpectralWaste, which may use slightly different training schedules and hardware.

For K3I-Cycling, material segmentation and plastic-type segmentation are reported in \Cref{tab:material_eval} and \Cref{tab:plastic_type_eval}, respectively. Additional qualitative examples for plastic-type are shown in \Cref{fig_4}. Material-segmentation examples are provided in the appendix.

\begingroup
\setlength{\tabcolsep}{3pt}
\begin{table*}[!t]
\centering\footnotesize
\caption{SpectralWaste: quantitative comparison of SegFormer (MiT-B0/B2), Swin-T (RGB), adapted Swin-T (HSI-$K$), BCAF, and Logit fusion across RGB resolutions and HSI slice counts.}
\label{tab:spectralwaste_eval}
\begin{tabular}{ll|cccccc|c|cccc}
\toprule
\multirow{2}{*}{Backbone} & \multirow{2}{*}{Modality} & \multicolumn{6}{c}{IoU (\%) $\uparrow$} & \multirow{2}{*}{mIoU (\%) $\uparrow$} & \multirow{2}{*}{Img./s $\uparrow$} & \multirow{2}{*}{Params (M) $\downarrow$} & \multirow{2}{*}{GFLOPs $\downarrow$} \\
 &   & Film & Basket & Card. & Tape & Filam. & Bag &  &  &  &  \\
\midrule
MiT-B0   & RGB-256     & 74.8 & 76.3 & 76.7 & 31.0 &52.0 & 62.0 &$ 62.1 \pm 1.6$  &  226.3 & 7.739 & 2.983 \\
MiT-B0   & RGB-1024    &  70.1& 80.0  & 73.8& 43.0& 74.2 & 60.5 & $66.9 \pm 1.1$  & 77.1 & 7.739 & 63.325\\
MiT-B2   & RGB-256     &76.9 & 81.1 & 72.2 & 39.8 & 67.8& 67.5 &$67.6  \pm 0.8$ & 134.4 & 41.866 &  13.817\\
MiT-B2   & RGB-1024    & 76.4  & 81.1 & 64.4& 46.7 &78.5& 66.6  & $ 69.0\pm 0.8$  & 30.6 &  41.866 & 279.458\\
Swin-T & RGB-256    & 76.1& 77.9 & 76.1 & 38.1& 59.2 &67.4 &$65.8 \pm 1.2$ & 141.0 & 32.176 & 9.862\\
Swin-T & RGB-512    & 78.2& 81.4 & 79.2 & 45.0 & 71.4& 71.6& $71.1 \pm 0.6$ & 134.6 & 32.176 & 35.571  \\
Swin-T & RGB-1024    & 75.3& 82.9 & 78.4 & 45.4 & 77.5& 70.2&\boldmath $\mathbf{71.6 \pm 0.3}$ & 60.4 & 32.176 & 138.683 \\
Swin-T & RGB-2048    & 69.1 & 82.7 & 69.2 & 46.6 & 76.4 &66.2 & $ 68.4\pm 0.8 $ & 15.0 & 32.176 & 546.316 \\
\midrule
Swin-T   & HSI-1     & 67.4 & 71.5 & 84.9 & 26.8 & 56.9& 57.7 &\boldmath $ \mathbf{60.9 \pm 0.2}$ & 141.0 & 32.176 & 9.906\\
Adapted Swin-T & HSI-3    & 68.3  & 66.9 & 86.8  & 24.5 & 55.7 & 56.1 & $ 59.7 \pm 0.7$ & 114.1 & 45.459 & 34.158 \\
Adapted Swin-T  & HSI-5    & 68.2  & 67.6 & 86.4  & 23.8 & 56.4 & 59.6 & $ 60.3 \pm 0.9$ & 118.8 & 45.417 & 54.343 \\
Adapted Swin-T  & HSI-7    & 66.3  & 64.7 & 83.5  & 24.2 & 58.1 & 57.1 & $ 59.0 \pm 1.5$ & 90.8 & 45.402 & 74.547 \\
Adapted Swin-T  & HSI-10   & 67.2  & 64.4 & 85.5  & 21.0 & 54.5 & 54.3 & $ 57.8 \pm 1.2$ & 67.9 & 45.394 & 104.902 \\
\midrule
CMX~\cite{zhang2023cmx}& RGB 256 + HSI-PCA3  & 71.7 & 71.6& 71.7 &27.8& 37.7 &59.4 & 56.6 & -- &11.193 &--\\
FusionSort~\cite{ali2025fusionsort} & RGB 256 + HSI-PCA3 &66.1 &72.5 &86.1 &24.7 &55.8 &56.6 &61.3 & -- & 6.285&--\\
HLRFF-Net~\cite{li2025hybrid} & RGB 256 + HSI-PCA3 & 75.4	&82.3	&83.5	&43.9	&73.2	&69.2 & 71.3 & --  & --  & -- \\ 
Logit Fusion & RGB-1024 + HSI-5    & 76.5& 82.2 & 83.9 & 47.2 & 74.9& 70.9 & $ 72.6\pm 0.8$ & 38.8 & 77.593 & 209.797\\
BCAF & RGB-256 + HSI-5    & 78.3 & 80.1 & 88.2 & 40.0 & 70.3& 69.7 & $ 71.1\pm 0.4$ & 54.4 & 101.296 &  78.284\\
BCAF & RGB-512 + HSI-5    & 80.3 & 84.7 & 90.1 & 47.2 & 76.9 & 73.2 & $ 75.4\pm 0.2$ & 54.9 & 101.296 & 119.007 \\
BCAF   & RGB-1024 + HSI-1    &78.8 & 84.6 & 87.7 &50.0& 80.9& 72.3& $75.7 \pm 0.5$ & 39.4 &  88.055 & 210.964 \\
BCAF & RGB-1024 + HSI-5    & 78.1 & 85.0 & 90.8&  50.0& 80.4 & 73.8 &\boldmath $  \mathbf{76.4\pm0.4 }$ & 31.4 & 101.296 & 282.176\\
\bottomrule
\end{tabular}
\end{table*}
\endgroup

\begingroup
\begin{table*}[!t]
\centering\footnotesize
\caption{K3I-Material (segmentation): quantitative comparison of Swin-T (RGB), adapted Swin-T (HSI-$K$), and BCAF across RGB resolutions and HSI slice counts.}
\label{tab:material_eval}
\begin{tabular}{llll|cccc|c}
\toprule
\multirow{2}{*}{Backbone} & \multirow{2}{*}{Modality} & \multicolumn{2}{c}{RGB} & \multicolumn{4}{c}{IoU (\%) $\uparrow$} & \multirow{2}{*}{mIoU (\%) $\uparrow$} \\
 &  & Origin Res & Input Res & other & plastic & paper & metal &  \\
\midrule
Swin-T & RGB & 256 & 256  & 15.7 & 76.5& 44.7  &7.1  &$ 36.0\pm 0.4$   \\
Swin-T & RGB & 256 & 512  & 25.4 & 80.2 & 50.9 & 16.3 &  $ 43.3\pm0.4 $  \\
Swin-T & RGB & 256 & 1024 & 24.6 & 81.2 & 53.4   & 18.6 & $ 44.4\pm 0.4$  \\
Swin-T & RGB & 256 & 2048 & 27.1 & 80.7 & 51.5   & 15.5 & $ 43.7 \pm 0.6$  \\
\midrule
Swin-T & RGB & 4096 & 512  & 27.2 & 80.2 & 51.5   & 23.8 & $ 45.7\pm0.5 $   \\
Swin-T & RGB & 4096 & 1024 & 33.8 & 85.3 & 61.5   & 33.4 & $53.5 \pm 0.6$  \\
Swin-T & RGB & 4096 & 2048 & 33.7  & 87.8 & 70.1   & 38.6 & \boldmath$ \mathbf{57.6\pm 0.4}$ \\
\midrule
Adapted Swin-T & HSI-1 & & &  10.1 & 83.9 & 72.7 &28.9 & $ 48.9\pm 0.6$   \\
Adapted Swin-T & HSI-3 & & &  18.2 &82.5  & 73.3 & 26.3 & \boldmath$ \mathbf{50.1\pm 0.9}$ \\
Adapted Swin-T & HSI-5 & & &  21.4 &82.4  & 73.4 & 19.8 & $ 49.2\pm 0.8$  \\
Adapted Swin-T & HSI-7 & & &  19.9 & 79.5  & 71.8 & 18.0 & $ 47.3\pm 1.3$  \\
Adapted Swin-T & HSI-10& & &  16.1 & 79.5  & 70.2 & 15.8 & $ 45.4\pm 0.8$  \\
\midrule
Logit Fusion& RGB+HSI-3 &  4096 & 1024   & 33.9  & 89.7 & 78.1 & 29.9 & $ 57.9\pm 1.8$  \\
BCAF & RGB+HSI-3 &  4096 & 1024   & 36.2  & 90.9 & 80.8 & 41.4 &\boldmath $ \mathbf{62.3\pm 1.1}$  \\
\bottomrule
\end{tabular}
\end{table*}
\endgroup

\begin{table*}[!t]
\setlength{\tabcolsep}{3pt}
\centering\footnotesize
\caption{K3I-Plastic (segmentation): quantitative comparison of Swin-T (RGB), adapted Swin-T (HSI-$K$), and BCAF across RGB resolutions and HSI slice counts.}
\label{tab:plastic_type_eval}
\begin{tabular}{llll|cccccccc|c}
\toprule
\multirow{2}{*}{Backbone} & \multirow{2}{*}{Modality} & \multicolumn{2}{c}{RGB} & \multicolumn{8}{c}{IoU (\%) $\uparrow$} & \multirow{2}{*}{mIoU (\%) $\uparrow$} \\
 &  & Origin Res & Input Res & no\_plastic & other\_plastic & HDPE & LDPE & PS & PP & PET & EPS &  \\
\midrule
Swin-T & RGB & 256 & 256  & 41.4 & 22.9 & 13.8 & 33.4 & 6.7 & 18.1 & 19.1 & 55.5 & $ 26.3\pm 0.6$  \\
Swin-T & RGB & 256 & 512  & 44.3 & 20.5 & 17.3 & 41.2 & 4.4  & 20.3 & 19.1 & 52.0 & $ 27.4\pm 1.1$ \\
Swin-T & RGB & 256 & 1024 & 51.5 & 24.5 & 16.3 & 40.6 & 9.0  & 25.8 & 22.3 & 62.6 & $ 31.6\pm 0.8$ \\
Swin-T & RGB & 256 & 2048 & 52.4 & 25.8 & 14.1 & 41.6 & 9.5  & 25.3 & 23.1 & 54.5 & $ 30.8\pm 0.5$ \\
\midrule
Swin-T & RGB & 4096 & 512  & 49.7 & 25.6 & 18.5 & 41.6 & 13.2 & 24.2 & 23.3 & 68.5 & $ 33.1\pm 0.6$ \\
Swin-T & RGB & 4096 & 1024 & 56.3 & 26.4 & 19.8 & 46.3 & 9.7  & 32.6 & 30.8 & 82.2 & $ 38.0\pm 0.1$ \\
Swin-T & RGB & 4096 & 2048 & 63.2 & 27.6 & 15.0 & 47.1 & 11.5 & 32.0 & 28.6 & 90.5 & \boldmath $ \mathbf{39.4\pm 0.2}$ \\
\midrule
Adapted Swin-T & HSI-1  &  & &  65.0 & 31.3 & 26.6 & 41.2 & 26.6& 36.9 & 50.6 & 71.6 &  $ 43.7\pm 1.1$ \\
Adapted Swin-T & HSI-3  &  &  & 68.2 & 34.1 & 47.0 & 65.5 & 67.3 & 60.8 & 71.0 & 85.3 & $ 62.4\pm 2.3$ \\
Adapted Swin-T & HSI-5  &  &  & 69.7 & 31.9 & 46.7 & 65.6 & 61.3 & 63.6 & 73.8 & 79.6 & $ 61.5\pm 1.7$ \\
Adapted Swin-T & HSI-7  &  &  & 69.5 & 35.8 & 56.6 & 68.0 & 65.0 & 62.9 & 74.9 & 82.6 & \boldmath $  \mathbf{64.4\pm 1.7}$ \\
Adapted Swin-T & HSI-10 &  &  & 68.4 & 36.9 & 54.0 & 66.3 & 65.8 & 63.6 & 74.3 & 85.1 & $ 64.3 \pm 0.8$ \\
\midrule
Logit Fusion & RGB+HSI-7 &  4096& 1024 & 69.7 & 31.6 & 45.6 & 65.6 & 46.8 & 63.5 & 55.3 & 85.8 &  $58.0  \pm  4.5$ \\
BCAF & RGB+HSI-7 &  4096& 1024 & 77.8 & 39.8 & 39.9 & 71.9 & 63.7 & 67.0 & 75.9 & 93.8 &\boldmath $\mathbf{ 66.2\pm 1.7}$\\

\bottomrule
\end{tabular}
\end{table*}

\subsection{RGB: effect of input resolution}
\label{subsec:RGB_effect_of_input_resolution}
\begin{figure}[ht]
    \centering
    \includegraphics[width=1.0\linewidth,trim={3.9cm 0.3cm 2.0cm 0.5cm},clip]{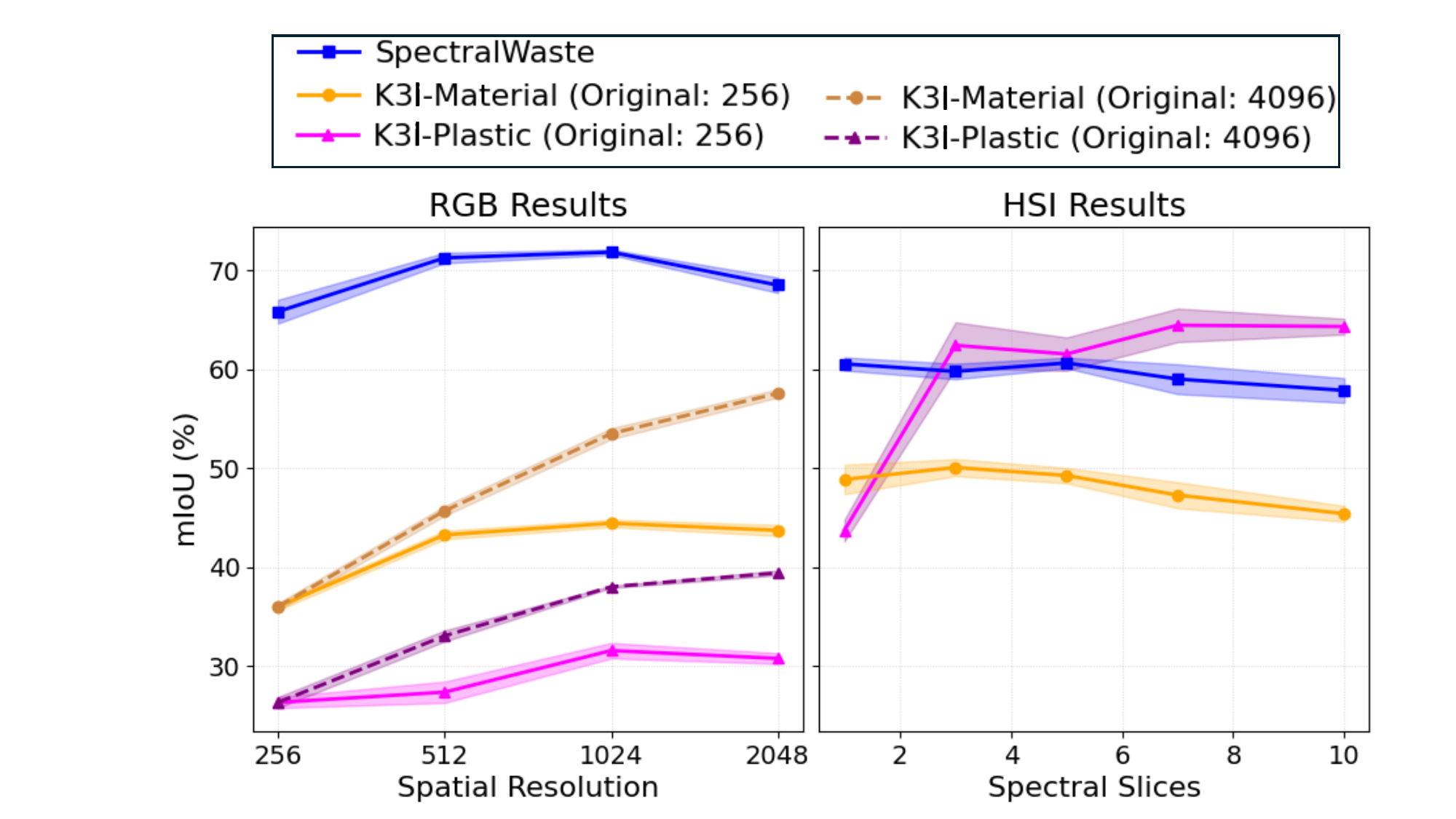}
    \caption{Effect of input resolution (left, RGB) and spectral slice count $K$ (right, HSI) on segmentation performance across SpectralWaste and K3I-Material/Plastic. Curves summarize the trends of \Cref{subsec:RGB_effect_of_input_resolution,subsec:HSI_effect_of_spectral_slices}. Higher mIoU is better.}
    \label{fig_5}
\end{figure}

We analyze the RGB backbone under the two resolution protocols defined in \Cref{subsec:Datasets}: (i) scaling-only (inputs upsampled but evaluated at $256{\times}256$, no new scene detail), and (ii) information-preserving (downsampling the native $4096{\times}4096$ acquisition to the target size, larger inputs retain more spatial detail).
Trends are visualized in \Cref{fig_5} (left).

\textbf{Scaling-only (SpectralWaste and K3I-Material/Plastic).}
Moderate upsampling ($256 \rightarrow 512 \rightarrow 1024$) improves segmentation by several percentage points, then degrades when pushed too far (2048). Because evaluation remains at $256{\times}256$, these gains reflect better optimization and more effective receptive-field/window utilization rather than additional spatial information. Excessive upscaling eventually hurts generalization. See \Cref{tab:spectralwaste_eval,tab:material_eval,tab:plastic_type_eval}. 

\textbf{Information-preserving (K3I-Material/Plastic).}
When working with native resolution images ($4096{\times}4096$), increasing the input size $256 \rightarrow 512 \rightarrow 1024 \rightarrow 2048$  yielding consistent gains. At the same nominal input sizes, the information-preserving protocol outperforms scaling-only, indicating that improvements are driven by genuinely retained scene detail (finer edges, small instances, textures) rather than scaling alone. We do not report $4096$ inputs due to computational constraints. See \Cref{tab:material_eval,tab:plastic_type_eval}.

\textbf{Backbone comparison (MiT vs.\ Swin).}
At an input resolution of 256 MiT-B2 outperforms Swin-T. When increasing the input size, swin’s shifted-window design benefits more from, whereas MiT’s spatial-reduction attention saturates earlier. See \Cref{tab:spectralwaste_eval}.

\subsection{HSI: effect of spectral slices}
\label{subsec:HSI_effect_of_spectral_slices}
We vary the number of spectral slices $K$ by grouping contiguous bands in the HSI backbone (\Cref{subsec:training_and_implementation_details}). We find that optimal $K$ selection is driven by task difficulty. Increasing $K$ exposes richer spectral structure, while decreasing $K$ denoises/regularizes.

\textbf{Texture-dominated (SpectralWaste, K3I-Material).}
Here we find that small to intermediate $K$ is sufficient. Collapsing the spectrum ($K{=}1$, 2D Swin without spectral attention) remains competitive for unimodal HSI. Large $K$ can even degrade segmentation performance due to amplified noise and limited need for fine spectral detail. The unimodal HSI-RGB gap is modest in these settings, indicating that texture/shape cues already carry most of the discriminative signal. See \Cref{tab:spectralwaste_eval,tab:material_eval}.

\textbf{Visually similar (K3I-Plastic).}
When material distinction is impossible using only visual cues, explicit spectral modeling is essential. Performance increases steadily with $K$ and peaks at moderate-to-large $K$ (around $7$, with $10$ close). Here, unimodal HSI with multi-slice spectral attention yields substantial mIoU gains over both unimodal RGB and HSI-1 (2D-only), confirming that polymer separation is governed by spectral signatures rather than texture or color. This shows the value of our HSI-adapted Swin, in contrast to approaches that rely on early spectral collapse (e.g., PCA into 2D processing), which underutilize HSI. Trends are visualized in \Cref{fig_5} (right), quantitative references appear in \Cref{tab:plastic_type_eval}.

\subsection{Fusion Results}
\label{subsec:Results_Fusion}
We compare BCAF against unimodal RGB/HSI baselines and a learned late-fusion baseline (logit fusion), all trained with identical backbones and schedules.

\textbf{Overall.}
Across datasets, BCAF consistently improves mIoU and per-class IoU over the strongest unimodal RGB/HSI baselines and over the learned late-fusion baseline. On SpectralWaste, BCAF achieves strong performance under the official split (see \Cref{tab:spectralwaste_eval}). Here the most directly comparable setting is RGB-256 + HSI-256, where all methods operate at the released 256$\times$256 inputs. In this matched-resolution regime, BCAF reaches 71.1\% mIoU. Our highest mIoU of 76.4\% is obtained with RGB upsampled to 1024$\times$1024 in the scaling-only regime (more pixels, same information), which increases GFLOPs. On K3I-Material/Plastic, BCAF improves both taxonomies. Gains are larger for K3I-Material, where BCAF really benefits from the complementary cues of RGB and HSI, and smaller but consistent for Plastic-type, where unimodal HSI is already strong (see \Cref{tab:material_eval,tab:plastic_type_eval}).

\textbf{Why BCAF beats logit fusion.}
Logit fusion aggregates decisions but lacks feature-level interactions. In contrast, BCAF performs bidirectional, local cross-attention between fine-grid RGB and coarse spectral slices, followed by gated fusion that modulates HSI contributions per channel and stage. This yields systematically higher, more stable gains than decision-level fusion at the same input sizes. On K3I-Plastic, where the mIoU gap between RGB and HSI is large (25 percentage points), logit fusion with unfrozen backbones allowed the weaker RGB stream to degrade strong HSI predictions, producing lower mIoU than HSI alone.

\begin{figure*}[!t]
    \centering
    \includegraphics[width=1.0\linewidth,trim={5.2cm 0.3cm 5.8cm 0.0cm},clip]{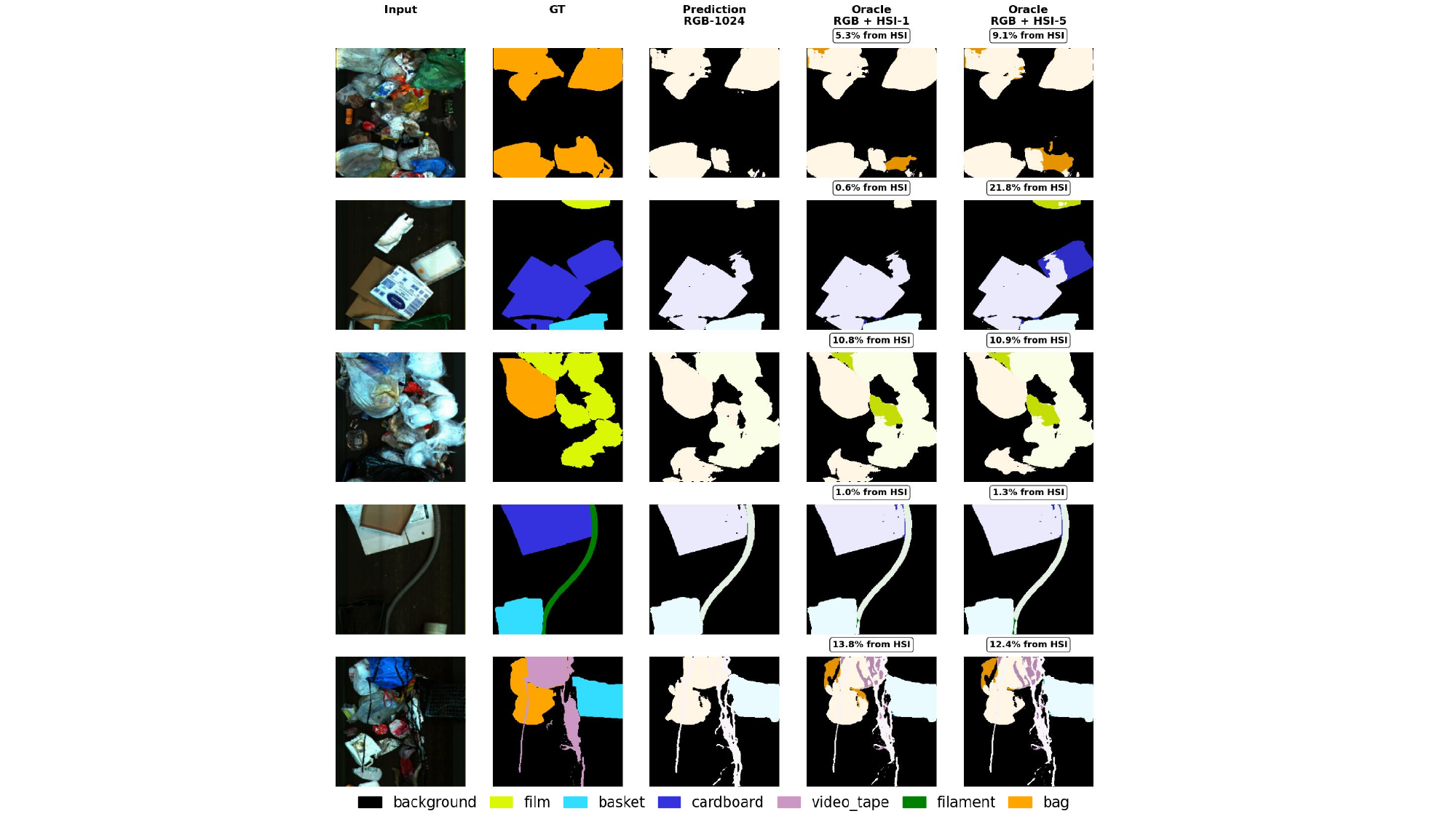}
    \caption{SpectralWaste: comparison of HSI-1 (K=1, spectral collapse) and HSI-5 (K=5, multi-slice). Oracle fusion (per-pixel best-of-modality using ground truth) shows a larger lift over RGB-1024 for HSI-5, evidencing stronger complementarity to RGB.}
    \label{fig_7}
\end{figure*}

\textbf{SpectralWaste: HSI-1 vs.\ HSI-5 in fusion.}
A central observation is that multi-slice HSI improves modality fusion even if the unimodal HSI-1 pipeline (spectral collapse, 2D processing) has a higher mIoU. With BCAF, RGB-$1024$+HSI-5 outperforms RGB-$1024$+HSI-1, because processing the $k=5$ slices contribute more complementary information to RGB. This can also be seen in the Oracle fusion, selecting, at each pixel, the correct prediction from RGB or HSI using ground truth (an upper bound on feature-level fusion without labels at test time). Here larger per-pixel lift over RGB can be seen HSI-5 than for HSI-1 (\Cref{fig_7}), indicating less redundancy and richer spectral cues. This complementarity yields the best semantic segmentation with RGB-$1024$+HSI-5 (\Cref{tab:spectralwaste_eval}).

\subsection{Qualitative Analysis}
\label{subsec:qualitative}
\begin{figure*}[!t]
    \centering
    \includegraphics[width=1.0\linewidth,trim={2.5cm 0.3cm 2.5cm 0.0cm},clip]{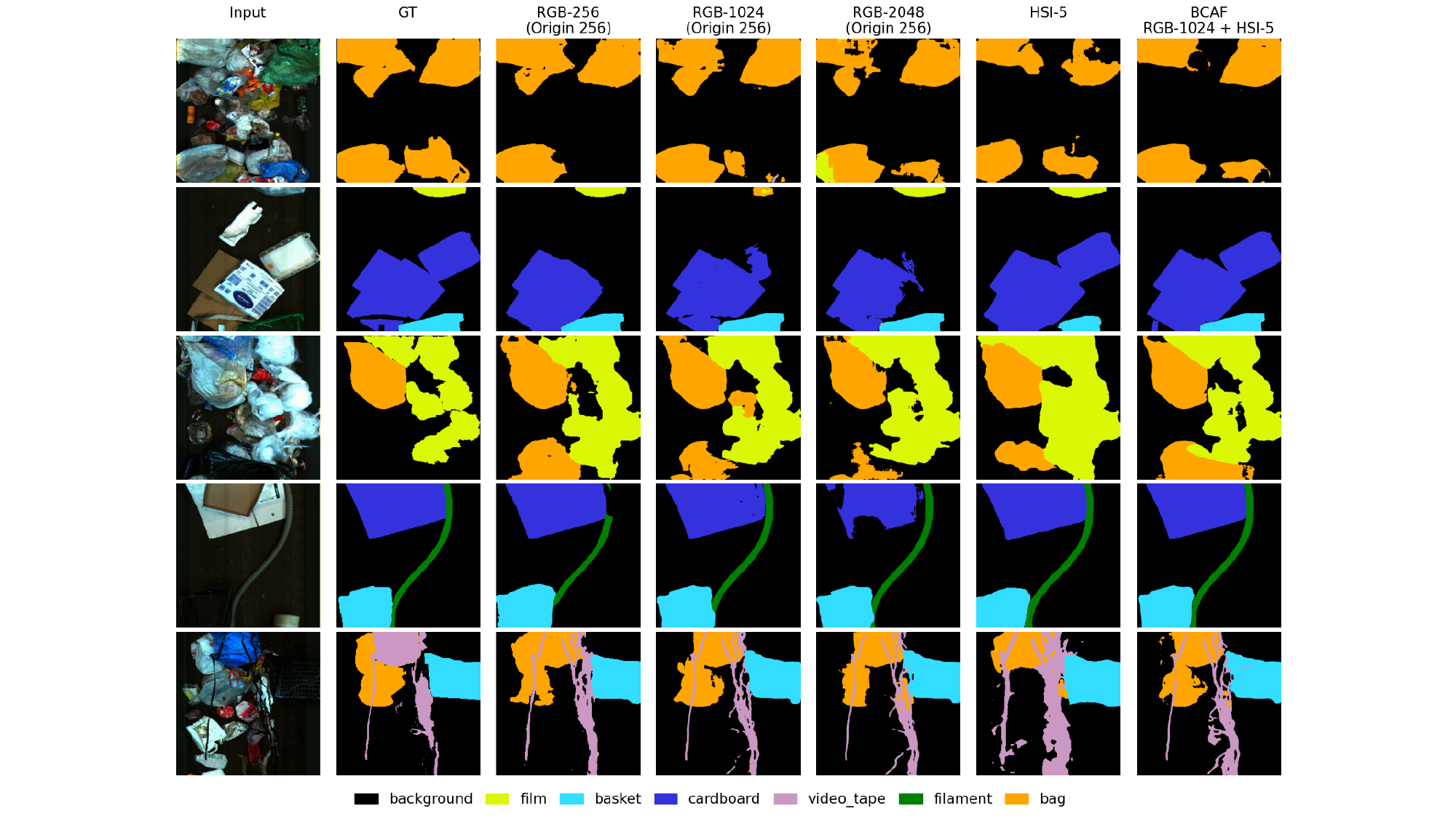}
    \caption{Qualitative results on SpectralWaste. Shown are Swin-T RGB at $256/1024/2048$, adapted Swin-T HSI with $K{=}5$ slices, and our best BCAF (RGB 1024 + HSI-5).}
    \label{fig_2}
\end{figure*}

\begin{figure*}[!t]
    \centering
    \includegraphics[width=1.0\linewidth,trim={0.9cm 0.3cm 1.3cm 0.0cm},clip]{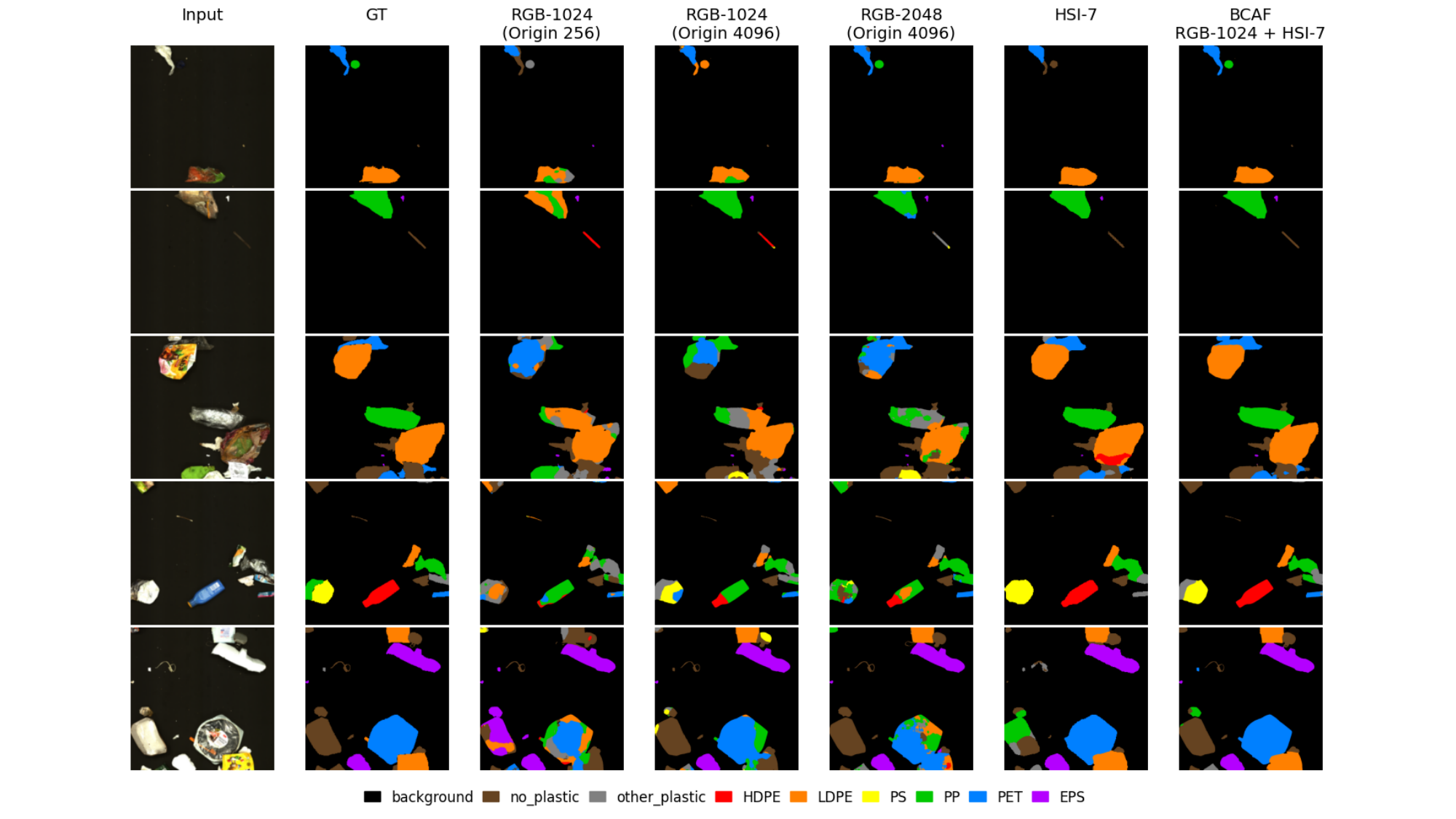}
    \caption{Qualitative results on K3I-Plastic (segmentation). Shown are Swin-T RGB at $1024$ and $2048$, adapted Swin-T HSI with $K{=}7$ slices, and our best BCAF (RGB 1024 + HSI-7).}
    \label{fig_4}
\end{figure*}

To complement the quantitative results, we qualitatively analyze typical success and failure cases on SpectralWaste and K3I-Plastic (Figs.~\ref{fig_2}, \ref{fig_4}, and Appendix examples).

\textbf{SpectralWaste.}
On SpectralWaste (Fig.~\ref{fig_2}), BCAF consistently sharpens object boundaries compared to both RGB-only and HSI-only baselines. It recovers structures that are primarily visible in HSI, such as the dark-blue cardboard region in row 2 that the RGB model misses, and it resolves overlaps where cardboard and filament or basket and filament intersect (row 4, green), where RGB fails to separate the classes cleanly. At the same time, BCAF preserves fine-grained spatial detail from RGB: in row 5, for example, the video-tape segmentation is noticeably slimmer and more accurate with BCAF than with HSI alone, reflecting the benefit of high-resolution RGB edges.

\textbf{K3I-Plastic.}
On K3I-Plastic (Fig.~\ref{fig_4}), we compare RGB models trained in the scaling-only regime (origin 256, upsampled; column 3) with models that retain native spatial information from the 4096$\times$4096 RGB images (column 4). Higher native resolution yields more complete and cleaner segmentations: for example, PP (green) is correctly recovered in row 2 only when using the high-resolution RGB input, and PET (blue) and no-plastic (grey) regions are more accurately delineated in row 5. The importance of HSI (column 6) becomes evident when comparing with RGB: in row 5, PET (blue) and LDPE (orange) are better separated spectrally than with RGB alone, and in row 4 the HDPE bottle is more clearly identified. However, HSI-only still misses very small structures (e.g., tiny “other plastic” fragments in grey), reflecting its lower spatial resolution. BCAF combines these strengths: in rows 1–2 it correctly classifies all objects and in rows 3–5 it recovers nearly all instances with correct labels while retaining sharp boundaries.

These qualitative observations align with the quantitative trends: fusion yields the largest gains when RGB and HSI provide complementary cues.

\subsection{Computation Analysis}
\label{subsec:Results_Computation}
We assess inference efficiency and complexity via throughput (images/s), parameters, and theoretical GFLOPs, using the protocol described in \Cref{subsec:Metric_and_Evaluation_Protocol}. Summary values appear in \Cref{tab:spectralwaste_eval}. 

\textbf{RGB scaling.}
As input size increases, GFLOPs grow quadratically while parameters remain constant. From $256\!\rightarrow\!512$, throughput is nearly unchanged (GPU underutilization at small sizes dominates). From $512\!\rightarrow\!1024$, throughput drops noticeably, and $2048$ is compute-heavy with a further decrease. Combined with the accuracy trends, $512$ yields an almost “free” boost in accuracy, and $1024$ offers a balanced operating point.

\textbf{HSI scaling.}
Increasing the spectral slice count $K$ adds approximately linear compute due to spectral self-attention. Throughput remains nearly flat up to small/intermediate $K$ (e.g., $K{\le}5$) and declines for larger $K$ (e.g., $7$-$10$) as spectral attention begins to dominate latency. 

\textbf{Fusion overhead (BCAF).}
BCAF computes cross-attention locally at each coarse HSI location, so that $r^2$ fine-grid RGB “children” interact with $K$ spectral slices. This leads to a score-computation cost per stage that scales as $\mathcal{O}\!\big(H_c W_c\, r^2 K\big)$, rather than $\mathcal{O}\!\big(r^2 K\, (H_c W_c)^2\big)$ for global all-to-all attention. In practice, the RGB and HSI backbones run efficiently in parallel, so the fusion overhead is modest. With RGB at $512$ and $K{=}5$ HSI slices, BCAF maintains real-time throughput while improving accuracy. Increasing RGB to $1024$ with $K{=}5$ reduces throughput as expected but remains viable for relaxed-latency settings.

\subsection{Ablation Study}
\label{subsec:Results_Ablation}
We ablate key components of the BCAF to validate design choices and quantify their impact under controlled settings. All runs share identical data preprocessing, augmentations, optimization as in \Cref{subsec:training_and_implementation_details}. 

We first ablate the HSI backbone in a unimodal setting on SpectralWaste at $256{\times}256$ with $K{=}5$ slices to isolate spectral design choices. We probe three questions: (i) how to embed raw bands into $K$ spectral slices (embedding), (ii) how to model spectral relations in the backbone (attention vs.\ convolution), and (iii) how to reduce the slice axis before the shared 2D decoder (spectral reduction). Results are in \Cref{tab:ablation_hsi_all}.

We then ablate fusion stage activation and directionality (RGB$\rightarrow$HSI, HSI$\rightarrow$RGB, bidirectional) and modality misalignment. We use RGB$=1024$ and HSI-5 on SpectralWaste, and additionally report K3I-Material/Plastic results with RGB$=1024$ and $K{=}3$ and $K{=}7$ respectively. Results are shown in \Cref{tab:ablation_fusion_big,tab:align}.

\begin{table*}[t]
\centering
\footnotesize
\setlength{\tabcolsep}{6pt}
\caption{HSI (unimodal) ablations on SpectralWaste: embedding variants, spectral modeling, and spectral reducer. All use $K{=}5$ unless noted. $\Delta$ is measured against the baseline row within each block.}
\label{tab:ablation_hsi_all}
\begin{tabular}{l l| c |c c c}
\toprule
Block & Variant &  mIoU (\%) $\uparrow$ &Img./s $\uparrow$ & Params (M) $\downarrow$& GFLOPs  $\downarrow$ \\
\midrule
\multirow[t]{5}{*}{\emph{Embedding}}
 & Grouped shared-kernel (baseline) &  \boldmath$ \mathbf{60.3 \pm 0.9}$ & 118.8 & 45.417 & 54.343\\
\emph{variants} & Single large projection  &   $56.7 \pm 1.5 $ & 117.9 & 47.076 & 60.005  \\
 & Unshared per-group kernels     &  $ 56.1 \pm 2.5$ & 120.4 & 45.694 & 54.343  \\
 & PCA-5                             &$ 57.6 \pm 1.6 $& 123.1 & 45.356 & 52.959   \\
 & SavGol + PCA-5                    &  $53.9 \pm 1.5$ & 123.5 &45.356& 52.959   \\
\cmidrule(l){2-6}
\multirow[t]{2}{*}{\emph{Spectral modeling}}
 & Spectral attention (baseline) &   \boldmath$ \mathbf{60.3 \pm 0.9}$ & 118.8 & 45.417 & 54.343\\
\emph{in backbone} & 1D Conv    &  $ 58.1 \pm 1.1$& 131.1 & 41.084& 49.790  \\
\cmidrule(l){2-6}
\multirow[t]{2}{*}{\emph{Spectral reducer}}
 & Spectral SE (baseline)             &  \boldmath$ \mathbf{60.3 \pm 0.9}$ & 118.8 & 45.417 & 54.343\\
 \emph{before decoder}& Learnable Weighted Reducer & $60.0 \pm 0.8$  & 121.3 & 45.220 & 54.342 \\
\bottomrule
\end{tabular}
\end{table*}

\textbf{HSI backbone: embedding.}
Our baseline (grouped shared-kernel) preserves spectral structure by grouping contiguous bands ($R_G{=}45$) and applying a single shared $4{\times}4{\times}R_G\!\to C$ convolution per group, yielding $K$-slices. We compare (1) a single large projection that uses one $4{\times}4{\times}S$ projection over all $224$ bands to directly produce $K{=}5$ tokens per location (no grouping, no weight sharing), (2) unshared per-group kernels with the same grouping but separate $4{\times}4{\times}R_G$ kernels per group (no sharing), (3) PCA-5 that collapses $224\!\to\!5$ components globally before a $4{\times}4{\times}1$ patch-embed per component, and (4) SavGol + PCA-5 that applies Savitzky-Golay smoothing \cite{Schafer2011SG} prior to PCA. 

We find that preserving spectral locality with shared weights provides the best accuracy. In contrast, the single large projection and unshared per-group variants do not improve accuracy, while PCA-based variants consistently underperform, indicating that early spectral collapse discards discriminative narrowband structure needed downstream.

\textbf{HSI backbone: spectral modeling.}
At each spatial location, the baseline applies position-wise spectral self-attention over the $K$ slices with learnable spectral positional encodings (content-adaptive mixing). We replace this with a 1D Conv along the slice axis (fixed local spectral receptive field) to test whether explicit attention is necessary. Replacing attention with 1D convolution yields a mIoU drop, suggesting that fixed kernels underfit slice correlations and cannot adapt mixing to class- and instance-specific spectral patterns.

\textbf{HSI backbone: spectral reducer (pre-decoder).}
Before the shared 2D decoder, the baseline spectral SE performs input-adaptive per-slice and per-channel gating followed by a weighted sum over $K$ (\Cref{subsec:Spectral_SE_Pooling_Module}). We compare a learnable weighted reducer that uses a global, input-independent vector of $K$ weights (shared across all channels). Spectral SE yields slightly higher mIoU, indicating that conditioning the spectral slice weights on features improves robustness to noise and class-dependent spectral variability at negligible overhead.

\textbf{Fusion: BCAF stage-wise effect.}
Table~\ref{tab:ablation_fusion_big} ablates the fusion stage and directionality.
Across all three datasets, bidirectional cross‑attention outperforms both RGB$\rightarrow$HSI‑only and HSI$\rightarrow$RGB‑only variants, confirming the importance of exchanging information in both directions rather than treating one modality purely as query or key/value. Fusing at all-stages performs best, confirming the value of multi-scale, localized interactions. Fusing only at the last stage (with RGB used via the head skip connection elsewhere) is close on SpectralWaste but lags on K3I-Cycling. Especially mIoU drops on K3I‑Plastic, suggesting that early-stage cues aid alignment in more diverse scenes. Fusing only at the first stage consistently reduces mIoU. Unidirectional variants underperform bidirectional.

\begin{table*}[t]
\centering
\footnotesize
\setlength{\tabcolsep}{2pt}
\caption{Fusion ablations (RGB 1024). Bidirectional refers to both RGB$\rightarrow$HSI and HSI$\rightarrow$RGB within each fusion block.}
\label{tab:ablation_fusion_big}
\begin{tabular}{lcc|ccc|cccc}
\toprule
Config & Stages & Direction & SpectralWaste mIoU (\%) & K3I-Material mIoU (\%) & K3I-Plastic mIoU (\%) &Img./s $\uparrow$ & Params (M) $\downarrow$& GFLOPs  $\downarrow$\\
\midrule
Early only & 1   & bidirectional & $73.2 \pm 0.3$ & $59.5 \pm 0.8$ & $59.1 \pm 1.1$  & 35.4& 82.477& 228.937 \\
Late only  & 4   & bidirectional & $76.3 \pm 0.5$ & $58.9 \pm 0.7 $ & $ 61.9 \pm 1.5$  & 38.6& 96.579& 225.979\\
All stages & 1-4& bidirectional & \boldmath $\mathbf{76.4\pm0.4}$ & \boldmath $\mathbf{61.8 \pm 1.2}$ & \boldmath $ \mathbf{66.2\pm 1.7}$ & 31.4 & 101.296 & 282.176\\
All stages & 1-4& RGB$\rightarrow$HSI & $76.0 \pm 0.6$ & $60.4 \pm 1.5$ & $61.5 \pm 0.7$  & 32.6 & 98.156 & 228.908 \\
All stages & 1-4& HSI$\rightarrow$RGB & $76.3 \pm 0.3 $ & $\ 60.4 \pm 0.8$ & $60.3 \pm 1.5$  & 33.1 & 98.156 &  268.166\\
\bottomrule
\end{tabular}
\end{table*}

\textbf{Fusion: Modality misalignment.}
 We assess robustness to cross-modal registration errors by shifting HSI only while keeping RGB and labels fixed.
 All models are trained on aligned data, results can be seen in \Cref{tab:align}). Across datasets, BCAF is more robust than late logit fusion for small‑to‑moderate shifts, indicating that localized, multi‑scale cross‑attention buffers modest misregistrations. The degradation pattern mirrors modality information content: SpectralWaste changes little under HSI shifts (RGB‑dominated), K3I‑Material degrades moderately (balanced cues), and K3I‑Plastic is most sensitive (HSI‑dominated). At extreme shifts, when cross‑modal overlap largely disappears, fusion benefits diminish and the gap to late fusion narrows.

\begin{table*}[t]
\centering\footnotesize
\caption{Absolute mIoU (\%) under HSI-only spatial shifts (pixels on the HSI grid). RGB and labels remain fixed, while HSI is shifted by $(dx,dy)$ with zero padding. All models are trained on aligned data.}
\label{tab:align}
\begin{tabular}{l|c|cccccc}
\toprule
Model / Dataset & (0,0) mIoU & (2,2) & (4,4) & (8,8) & (16,16) & (32,32) & (64,64) \\
\midrule
Logit fusion (SpectralWaste)  & $72.6 \pm 0.8$ &  $72.7 \pm 0.8$ & $72.6\pm0.8$ & $72.2 \pm 0.6$ &  $69.5 \pm 1.2$ & $58.8 \pm 2.7$ &   $36.2\pm 4.9$ \\
BCAF (SpectralWaste)          & $76.4 \pm 0.4$ &  $76.3 \pm 0.5$ & $76.3 \pm 0.4$ & $76.0 \pm 0.3$ & $75.3 \pm 0.4$ & $73.4 \pm 0.6$ &   $69.1 \pm 1.3$ \\
\midrule
Logit fusion (K3I-Material)  &$57.9 \pm 0.2$&  $57.5 \pm 0.4$ &  $55.8 \pm 0.8$ &  $ 50.7 \pm 1.3$  &  $ 42.7 \pm  1.2$ &  $37.4 \pm 1.4$ & $36.6 \pm 1.1$  \\
BCAF (K3I-Material) &   $ 62.3 \pm 1.1$ & $62.2 \pm 1.2$ & $ 61.1 \pm 1.3$ &   $57.9 \pm 1.2$ & $50.6 \pm 1.3 $ & $ 45.0 \pm 1.41$ &   $ 44.8 \pm 1.4$   \\
\midrule
Logit fusion (K3I-Plastic)    & $58.0  \pm  4.5$     & $57.6  \pm  4.6$   & $55.6  \pm  4.2$    & $49.6  \pm  2.0$    &$38.3  \pm  1.3$     & $26.2  \pm  3.7$     & $21.2  \pm  4.2$   \\
BCAF (K3I-Plastic)  &         $66.2\pm 1.7$ &  $66.0 \pm 1.8$ & $63.8 \pm 1.8$ & $58.0 \pm 2.0$ &  $45.0 \pm 1.1$ & $ 27.0 \pm 1.0$ & $18.5 \pm 1.5$ \\
\bottomrule
\end{tabular}
\end{table*}

\textbf{Fusion: Training protocol.}
Initializing BCAF’s RGB and HSI backbones from the best unimodal checkpoints, then fine-tuning the full fusion model end-to-end, yields more stable convergence and higher mIoU than starting from ImageNet-initialized backbones ($55.9 \pm 1.8$ on SpectralWaste), with no architectural changes.

\section{Conclusion}
\label{sec:conclusion}
We addressed pixel-accurate, real-time waste sorting by fusing high-resolution RGB with lower-resolution HSI without collapsing spectra or eroding spatial detail. We introduced Bidirectional Cross-Attention Fusion (BCAF), which (i) adapts Swin Transformer to HSI via grouped spectral tokenization and factorized spatial-spectral attention, (ii) aligns fine-grid RGB and coarse-grid HSI features through localized, bidirectional cross-attention across multiple scales, and (iii) employs spectral SE pooling and gated fusion to share a single, lightweight decoder across RGB, HSI, and fused inputs.

Across SpectralWaste and our industrial K3I-Cycling datasets, three trends stand out. (1) Scaling the RGB input improves segmentation in both the scaling-only regime (more pixels, same information on SpectralWaste) and the information-preserving regime (more pixels, more information on K3I), with the biggest gains appearing when larger inputs preserve native high-resolution detail. (2) Preserving and attending along the spectral axis in our adapted HSI backbone provides genuinely complementary cues beyond 2D backbones and it clearly excels when spectral signatures dominate (e.g., plastic-type discrimination). (3) Building on these strengths, BCAF couples fine-grid RGB structure with multi-slice HSI features via localized, bidirectional cross-attention, consistently surpassing unimodal baselines and learned logit-level fusion and achieving strong performance on SpectralWaste at practical throughput.

A limitation of BCAF is that it assumes paired, well‑registered RGB-HSI acquisition: localized multi‑scale cross‑attention tolerates only small shifts, while severe spatial or temporal registration errors noticeably degrades fusion. The dual‑backbone design increases parameters and compute. Performance is further sensitive to the HSI slice count $K$: small $K$ can underuse spectral detail on fine‑grained tasks, whereas large $K$ improves discrimination at added latency and potential noise ($K$ should be tuned per task). Finally, the approach presumes synchronized sensors and calibration, which may limit deployment where reliable pairing is difficult to guarantee.

Future work includes extending BCAF to other co‑registered RGB+X sensing modalities (e.g., multispectral, NIR/SWIR, thermal, depth/ToF). We also plan to deploy it on a real conveyor system to assess end‑to‑end performance under realistic operating conditions. In addition, given the abundance of unlabeled HSI data and the lack of established hyperspectral foundation models, we plan to investigate self‑supervised pretraining strategies for the HSI backbone, analogous to how ImageNet pretraining benefits the RGB backbone, with the aim of further improving robustness and segmentation accuracy.

In summary, BCAF preserves hyperspectral structure and aligns it with high-resolution RGB through multi-scale, bidirectional cross-attention, delivering accurate and efficient multimodal segmentation.

\section*{Acknowledgement}
We would like to thank Steffen Rüger and his team of the Fraunhofer Institute for Integrated Circuits IIS for their support in dataset annotation. We also thank Lobbe RSW GmbH (Iserlohn, Germany) for providing samples of lightweight packaging waste. 

\subsection*{Data availability}
SpectralWaste: Publicly available at \url{https://github.com/ferpb/spectralwaste-dataset} ~\citep{casao2024spectralwaste}. We used the dataset under the terms of the original repository.

K3I-Cycling: Proprietary dataset collected at Fraunhofer IOSB with support from Lobbe RSW GmbH. Not fully publicly released yet. RGB subset already available at \url{http://dx.doi.org/10.24406/fordatis/420}. We are currently working on publishing the HSI part; access can already be provided upon request.

\subsection*{Code availability}
The full source code for BCAF, including training and evaluation scripts, 
configuration files, and model checkpoints, is publicly available at:
\url{https://github.com/jonasvilhofunk/BCAF_2026} (code) and 
\url{https://huggingface.co/jonasvilhofunk/BCAF_2026} (model weights).

\subsection*{Funding}
Funding was provided by the Federal Ministry of Research, Technology and Space (BMFTR) under the funding reference 033KI201.

\subsection*{Declaration of competing interest}
The authors declare no competing financial or non-financial interests.

\subsection*{Declaration of generative AI and AI-assisted technologies in the writing process}
During the preparation of this work the authors used FhGenie (GPT5-based) in order to improve readability. After using this tool, the authors reviewed and edited the content as needed and take full responsibility for the content of the publication.

\bibliographystyle{elsarticle-num}
\bibliography{refs}  

@inproceedings{casao2024spectralwaste,
  title={SpectralWaste Dataset: Multimodal Data for Waste Sorting Automation},
  author={Casao, Sara and Pe{\~n}a, Fernando and Sabater, Alberto and Castill{\'o}n, Rosa and Su{\'a}rez, Dar{\'\i}o and Montijano, Eduardo and Murillo, Ana C},
  booktitle={2024 IEEE/RSJ International Conference on Intelligent Robots and Systems (IROS)},
  pages={5852--5858},
  year={2024},
  organization={IEEE}
}

@article{li2025hybrid,
  title={Hybrid long-range feature fusion network for multi-modal waste semantic segmentation},
  author={Li, Yangke and Zhang, Xinman},
  journal={Information Fusion},
  pages={103608},
  year={2025},
  publisher={Elsevier}
}

@article{ali2025fusionsort,
  title={FusionSort: Enhanced Cluttered Waste Segmentation with Advanced Decoding and Comprehensive Modality Optimization},
  author={Ali, Muhammad and AlSuwaidi, Omar Ali},
  journal={arXiv preprint arXiv:2508.19798},
  year={2025}
}

@ARTICLE{Maier2024,
  author={Maier, Georg and Gruna, Robin and Längle, Thomas and Beyerer, Jürgen},
  journal={IEEE Access}, 
  title={A Survey of the State of the Art in Sensor-Based Sorting Technology and Research}, 
  year={2024},
  volume={12},
  number={},
}

@article{Ahmad2025,
   title={A comprehensive survey for Hyperspectral Image Classification: The evolution from conventional to transformers and Mamba models},
   volume={644},
   journal={Neurocomputing},
   author={Ahmad, Muhammad and Distefano, Salvatore and Khan, Adil Mehmood and Mazzara, Manuel and Li, Chenyu and Li, Hao and Aryal, Jagannath and Ding, Yao and Vivone, Gemine and Hong, Danfeng},
   year={2025},
}

@article{Liu2021,
  author       = {Ze Liu and
                  Yutong Lin and
                  Yue Cao and
                  Han Hu and
                  Yixuan Wei and
                  Zheng Zhang and
                  Stephen Lin and
                  Baining Guo},
  title        = {Swin Transformer: Hierarchical Vision Transformer using Shifted Windows},
  journal      = {CoRR},
  volume       = {abs/2103.14030},
  year         = {2021},
}

@article{Xie2021,
  author       = {Enze Xie and
                  Wenhai Wang and
                  Zhiding Yu and
                  Anima Anandkumar and
                  Jos{\'{e}} M. {\'{A}}lvarez and
                  Ping Luo},
  title        = {SegFormer: Simple and Efficient Design for Semantic Segmentation with
                  Transformers},
  journal      = {CoRR},
  year         = {2021},
}

@article{Hong2021,
  author       = {Danfeng Hong and
                  Zhu Han and
                  Jing Yao and
                  Lianru Gao and
                  Bing Zhang and
                  Antonio Plaza and
                  Jocelyn Chanussot},
  title        = {SpectralFormer: Rethinking Hyperspectral Image Classification with
                  Transformers},
  journal      = {CoRR},
  year         = {2021},
}

@ARTICLE{Yang2022,
  author={Yang, Xiaofei and Cao, Weijia and Lu, Yao and Zhou, Yicong},
  journal={IEEE Transactions on Geoscience and Remote Sensing}, 
  title={Hyperspectral Image Transformer Classification Networks}, 
  year={2022},
  volume={60},
  number={},
}

@Article{He2021,
AUTHOR = {He, Xin and Chen, Yushi and Lin, Zhouhan},
TITLE = {Spatial-Spectral Transformer for Hyperspectral Image Classification},
JOURNAL = {Remote Sensing},
VOLUME = {13},
YEAR = {2021},
NUMBER = {3},
ARTICLE-NUMBER = {498},
}

@ARTICLE{Sun2022,
  author={Sun, Le and Zhao, Guangrui and Zheng, Yuhui and Wu, Zebin},
  journal={IEEE Transactions on Geoscience and Remote Sensing}, 
  title={Spectral–Spatial Feature Tokenization Transformer for Hyperspectral Image Classification}, 
  year={2022},
  volume={60},
  number={},
}

@article{Ji2025,
title = {Plastic waste identification based on multimodal feature selection and cross-modal Swin Transformer},
journal = {Waste Management},
volume = {192},
year = {2025},
author = {Tianchen Ji and Huaiying Fang and Rencheng Zhang and Jianhong Yang and Zhifeng Wang and Xin Wang},
}

@book{Chang2003,
  author    = {Chein{-}I Chang},
  title     = {Hyperspectral Imaging: Techniques for Spectral Detection and Classification},
  publisher = {Springer Science \& Business Media},
  year      = {2003}
}

@book{Burns2007,
  title     = {Handbook of Near-Infrared Analysis},
  editor    = {Burns, David A. and Ciurczak, Emil W.},
  edition   = {3rd},
  year      = {2007},
  publisher = {CRC Press},
}

@article{senanayake2025automated,
  title={Automated Electro-construction waste Sorting: Computer vision for part-level segmentation},
  author={Senanayake, Aseni and Arashpour, Mehrdad},
  journal={Waste Management},
  volume={203},
  pages={114883},
  year={2025},
  publisher={Elsevier}
}

@article{paszke2016enet,
  title={Enet: A deep neural network architecture for real-time semantic segmentation},
  author={Paszke, Adam and Chaurasia, Abhishek and Kim, Sangpil and Culurciello, Eugenio},
  journal={arXiv preprint arXiv:1606.02147},
  year={2016}
}

@inproceedings{zhao2018icnet,
  title={Icnet for real-time semantic segmentation on high-resolution images},
  author={Zhao, Hengshuang and Qi, Xiaojuan and Shen, Xiaoyong and Shi, Jianping and Jia, Jiaya},
  booktitle={Proceedings of the European conference on computer vision (ECCV)},
  pages={405--420},
  year={2018}
}

@inproceedings{hu2018squeeze,
  title={Squeeze-and-excitation networks},
  author={Hu, Jie and Shen, Li and Sun, Gang},
  booktitle={Proceedings of the IEEE conference on computer vision and pattern recognition},
  pages={7132--7141},
  year={2018}
}

@inproceedings{shi2016real,
  title={Real-time single image and video super-resolution using an efficient sub-pixel convolutional neural network},
  author={Shi, Wenzhe and Caballero, Jose and Husz{\'a}r, Ferenc and Totz, Johannes and Aitken, Andrew P and Bishop, Rob and Rueckert, Daniel and Wang, Zehan},
  booktitle={Proceedings of the IEEE conference on computer vision and pattern recognition},
  pages={1874--1883},
  year={2016}
}

@misc{rw2019timm,
  author = {Ross Wightman},
  title = {PyTorch Image Models},
  year = {2019},
  publisher = {GitHub},
  journal = {GitHub repository},
  doi = {10.5281/zenodo.4414861},
  howpublished = {\url{https://github.com/rwightman/pytorch-image-models}}
}

@article{Schafer2011SG,
  author  = {Schafer, Ronald W.},
  title   = {What Is a Savitzky--Golay Filter?},
  journal = {IEEE Signal Processing Magazine},
  year    = {2011},
  volume  = {28},
  number  = {4},
  pages   = {111--117}
}

@article{zhang2023cmx,
  title={CMX: Cross-modal fusion for RGB-X semantic segmentation with transformers},
  author={Zhang, Jiaming and Liu, Huayao and Yang, Kailun and Hu, Xinxin and Liu, Ruiping and Stiefelhagen, Rainer},
  journal={IEEE Transactions on intelligent transportation systems},
  volume={24},
  number={12},
  pages={14679--14694},
  year={2023},
  publisher={IEEE}
}

@article{zhou2022canet,
  title={CANet: Co-attention network for RGB-D semantic segmentation},
  author={Zhou, Hao and Qi, Lu and Huang, Hai and Yang, Xu and Wan, Zhaoliang and Wen, Xianglong},
  journal={Pattern Recognition},
  volume={124},
  pages={108468},
  year={2022},
  publisher={Elsevier}
}

@misc{european2025,
  author       = {{European Parliament}},
  title        = {How to reduce plastic waste: EU measures explained},
  howpublished = {\url{https://www.europarl.europa.eu/topics/en/article/20180830STO11347/how-to-reduce-plastic-waste-eu-measures-explained#plastic-packaging-waste-10}},
  note         = {Accessed: 10 Dec 2025},
}

@inproceedings{bihler2023multi,
  title={Multi-sensor data fusion using deep learning for bulky waste image classification},
  author={Bihler, Manuel and Roming, Lukas and Jiang, Yifan and Afifi, Ahmed J and Aderhold, Jochen and {\v{C}}ibirait{\.e}-Lukenskien{\.e}, Dovil{\.e} and Lorenz, Sandra and Gloaguen, Richard and Gruna, Robin and Heizmann, Michael},
  booktitle={Automated Visual Inspection and Machine Vision V},
  volume={12623},
  pages={69--82},
  year={2023},
  organization={SPIE}
}

@inproceedings{long2015fully,
  title={Fully convolutional networks for semantic segmentation},
  author={Long, Jonathan and Shelhamer, Evan and Darrell, Trevor},
  booktitle={Proceedings of the IEEE conference on computer vision and pattern recognition},
  pages={3431--3440},
  year={2015}
}

@article{chen2017deeplab,
  title={Deeplab: Semantic image segmentation with deep convolutional nets, atrous convolution, and fully connected crfs},
  author={Chen, Liang-Chieh and Papandreou, George and Kokkinos, Iasonas and Murphy, Kevin and Yuille, Alan L},
  journal={IEEE transactions on pattern analysis and machine intelligence},
  volume={40},
  number={4},
  pages={834--848},
  year={2017},
  publisher={IEEE}
}

@inproceedings{Ronneberger2015,
author = {Ronneberger, Olaf and Fischer, Philipp and Brox, Thomas},
year = {2015},
booktitle ={Medical Image Computing and Computer-Assisted Intervention (MICCAI)},
title = {U-Net: Convolutional Networks for Biomedical Image Segmentation},
volume = {9351},
journal = {LNCS},
}

@article{dosovitskiy2021,
  title={An Image is Worth 16x16 Words: Transformers for Image Recognition at Scale},
  author={Dosovitskiy, Alexey and Beyer, Lucas and Kolesnikov, Alexander and Weissenborn, Dirk and Zhai, Xiaohua and Unterthiner, Thomas and Dehghani, Mostafa and Minderer, Matthias and Heigold, Georg and Gelly, Sylvain and Uszkoreit, Jakob and Houlsby, Neil},
  journal={International Conference on Learning Representations},
  year={2021},
}

\section*{Appendix}

\subsection*{Dataset Class Spectra}
\begin{figure*}[t]
    \centering
    \includegraphics[width=1.0\linewidth,trim={0.5cm 0.0cm 12.5cm 0.0cm},clip]{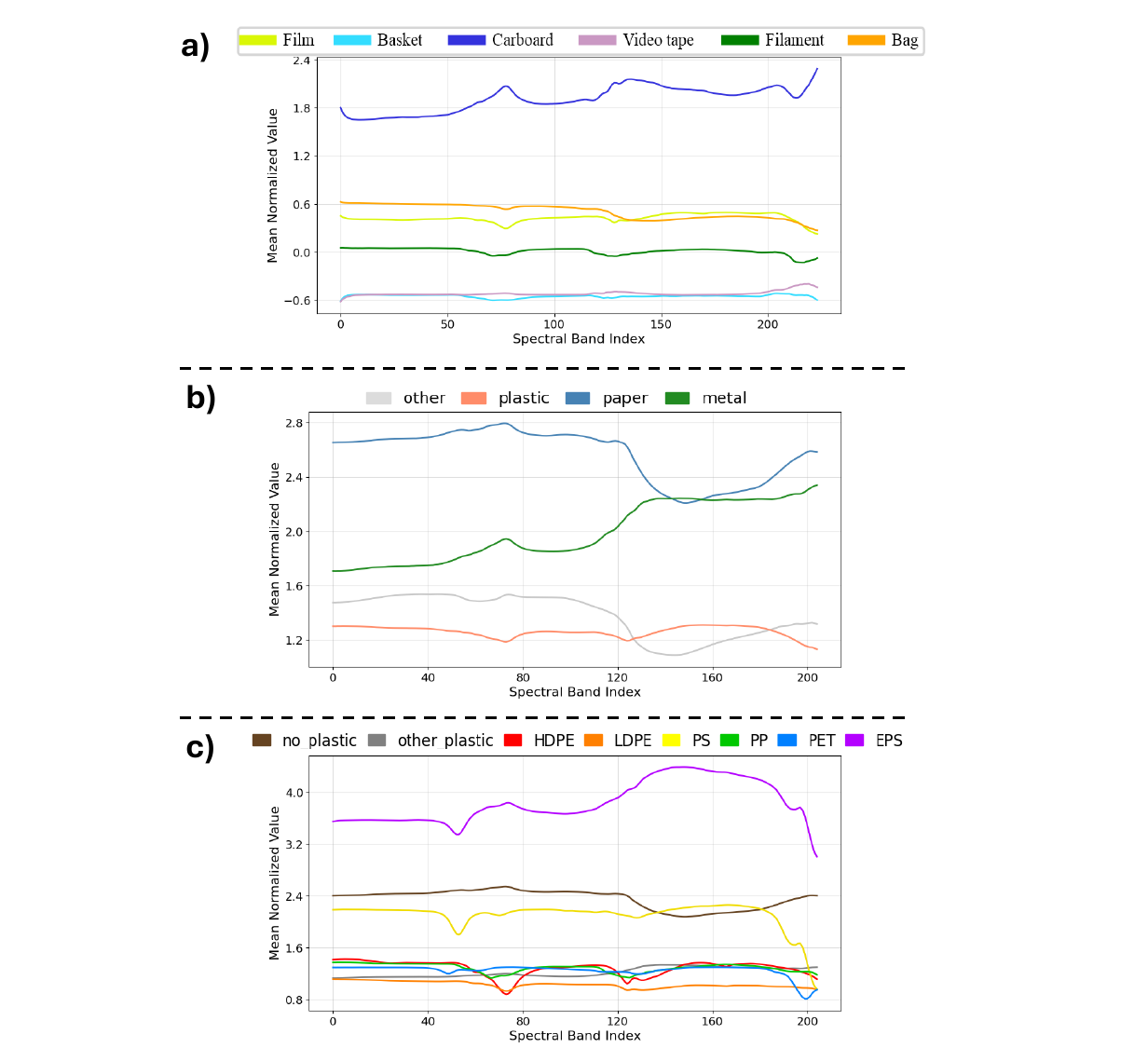}
    \caption{Mean normalized spectra per class for (a) SpectralWaste, (b) K3I-Material, and (c) K3I-Plastic.}
    \label{fig:spectra}
\end{figure*}

We show the mean normalized spectra for each class in all datasets. First, the HSI data is normalized. Then, for each class, all corresponding pixels are aggregated and the mean spectrum over channels is computed and plotted (see \Cref{fig:spectra}).

\subsection*{Semantic segmentation on K3I-Material}
Additional qualitative segmentation results on K3I-Material. We compare Swin-T RGB at 1024 and 2048, adapted Swin-T HSI with $K{=}3$ spectral slices, and BCAF (RGB-1024 + HSI-3). Relative to SpectralWaste (RGB at $256{\times}256$), the native-resolution regime on K3I-Material yields clear gains: downsampling from the $4096{\times}4096$ acquisition to 2048 improves performance, whereas pure upscaling from $256{\times}256$ degrades. BCAF improvements are especially visible for the paper/cardboard class. These observations align with the quantitative gains in \Cref{sec:results}.

\begin{figure*}[t]
    \centering
    \includegraphics[width=1.0\linewidth,trim={2.5cm 0.3cm 2.5cm 0.0cm},clip]{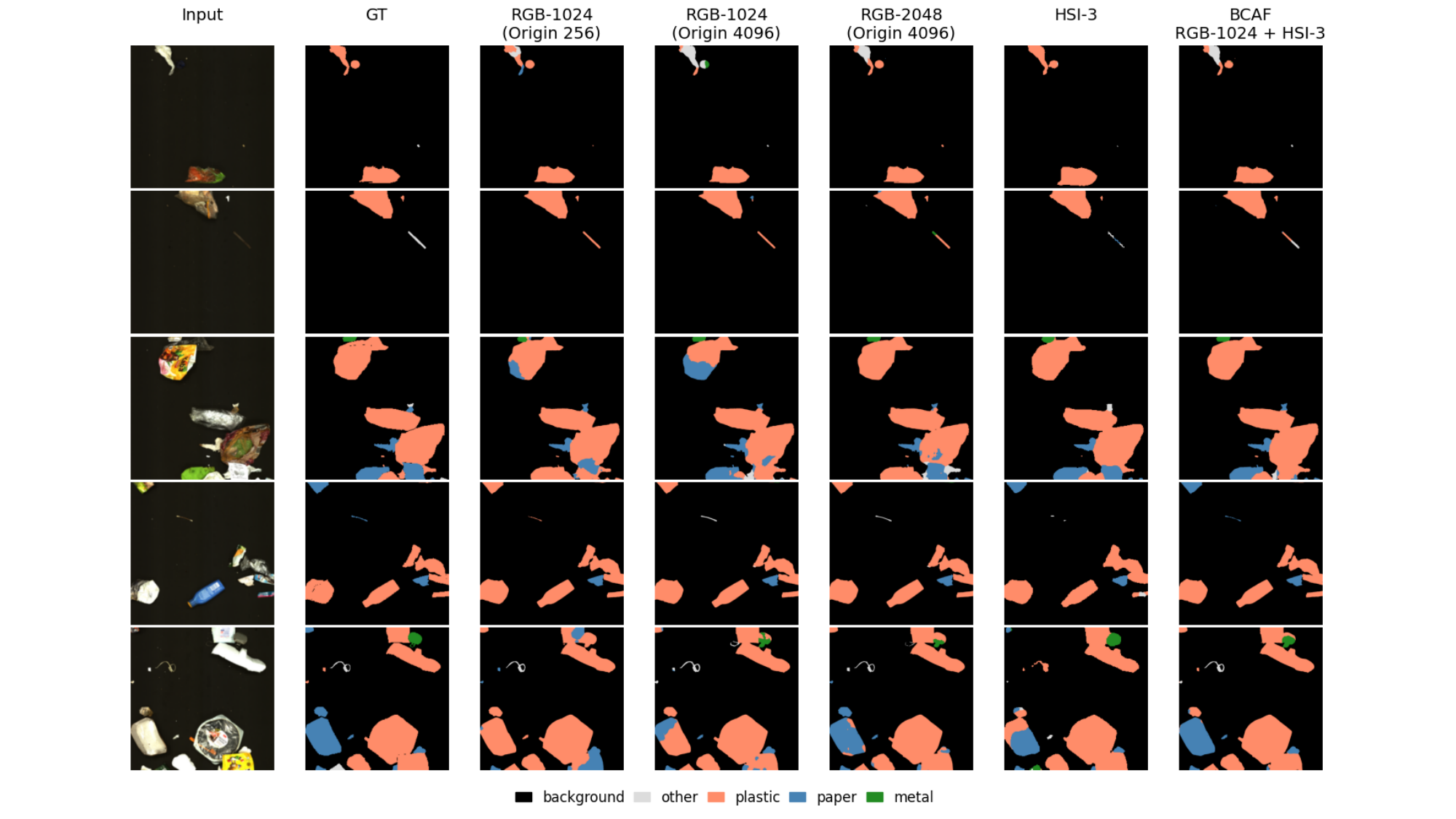}
    \caption{K3I-Material qualitative results. Shown are Swin-T RGB at 1024 and 2048, adapted Swin-T HSI-3, and BCAF (RGB 1024 + HSI-3).}
    \label{fig_3}
\end{figure*}

\subsection*{Visualization of Swin-T backbone feature activations}
We visualize Swin-T backbone activations across stages for input resolutions 256 vs 1024 by computing per-stage heatmaps. For each stage output $\mathbf{F_\mathrm{rgb}}\in\mathbb{R}^{H\times W\times C}$, we average over channels to obtain a single spatial map $\mathbf{A}=\mathrm{mean}_{c}(\mathbf{F_\mathrm{rgb}})\in\mathbb{R}^{H\times W}$. We apply min-max normalization to [0,1] and render with a magma colormap. 

\begin{figure}[t]
    \centering
    \includegraphics[width=1.0\linewidth,trim={6.5cm 0.0cm 18.0cm 0.0cm},clip]{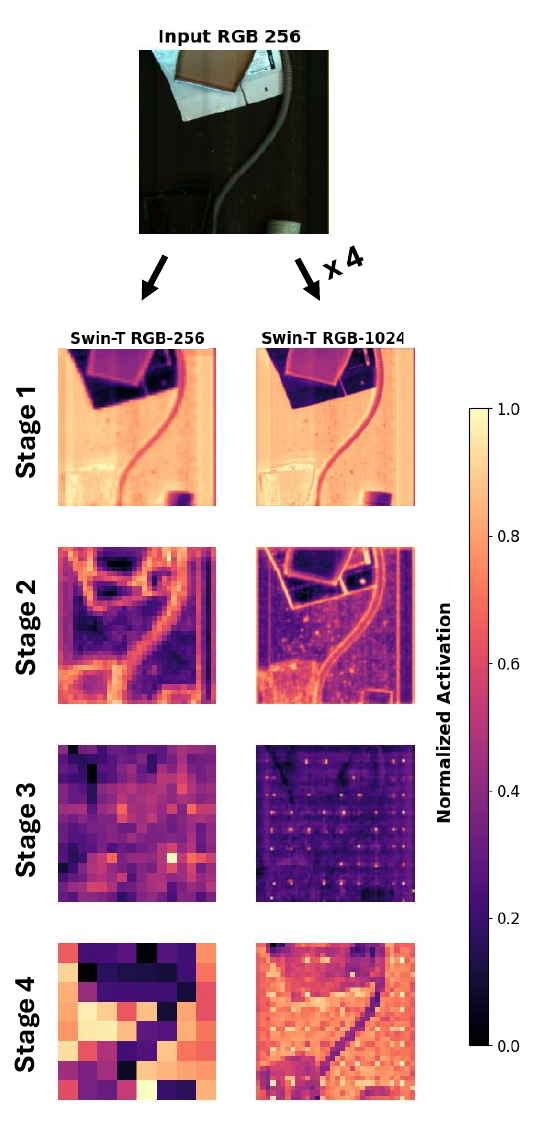}
     \caption{Swin-T feature activations across stages for input resolutions 256 vs 1024. Activations are mean over channels, normalized to [0,1].}
    \label{fig_6}
\end{figure}


\end{document}